\pdfoutput=1
\documentclass[acmlarge]{acmart}
\usepackage{svg}
\usepackage{subfigure}
\usepackage{algorithm}
\usepackage[noend]{algorithmic}
\usepackage{amsmath}
\usepackage{multirow}
\usepackage{bm}
\usepackage{diagbox} 
\usepackage{color,soul}

\AtBeginDocument{%
  \providecommand\BibTeX{{%
    \normalfont B\kern-0.5em{\scshape i\kern-0.25em b}\kern-0.8em\TeX}}}

\setcopyright{acmcopyright}
\copyrightyear{xx}
\acmYear{xx}
\acmDOI{xxxx}

\acmJournal{TIST}
\acmVolume{1}
\acmNumber{1}
\acmArticle{1}
\acmMonth{1}

\begin{document}

\title{Phased Flight Trajectory Prediction with Deep Learning}


\author{Kai Zhang}
\email{kaz321@lehigh.edu}
\affiliation{%
  \institution{Lehigh University}
}

\author{Bowen Chen}
\email{chenb2@my.erau.edu}
\affiliation{%
  \institution{Embry-Riddle Aeronautical University}
}






\renewcommand{\shortauthors}{Zhang and Chen.}

\begin{abstract}
The unprecedented increase of commercial airlines and private jets over the next ten years presents a challenge for air traffic control. Precise flight trajectory prediction is of great significance in air transportation management, which contributes to the decision-making for safe and orderly flights. Existing research and application mainly focus on the sequence generation based on historical trajectories, while the aircraft-aircraft interactions in crowded airspace especially the airspaces near busy airports have been largely ignored. On the other hand, there are distinct characteristics of aerodynamics for different flight phases, and the trajectory may be affected by various uncertainties such as weather and advisories from air traffic controllers. However, there is no literature fully considers all these issues. Therefore, we proposed a phased flight trajectory prediction framework. Multi-source and multi-modal datasets have been analyzed and mined using variants of recurrent neural network (RNN) mixture. To be specific, we first introduce spatio-temporal graphs into the low-altitude airway prediction problem, and the motion constraints of an aircraft are embedded to the inference process for reliable forecasting results. In the en-route phase, the dual-attention mechanism is employed to adaptively extract much more important features from overall datasets to learn the hidden patterns in dynamical environments.
The experimental results demonstrate our proposed framework can outperform state-of-the-art methods for flight trajectory prediction for large passenger/transport airplanes. 
\end{abstract}


\maketitle

\section{Introduction} \label{sec:introduction}
Worldwide, there is a dramatic increase in air traffic demand, which potentially results in more congested airspace and flight delay. To advance the aerial mobility and safety, some state agencies are developing air traffic management (ATM) modernization programs, for example, the Federal Aviation Administration in the United States developed Next Generation Air Transportation System (NextGen) to estimate sector capacity and enhance decision support. In Europe, EUROCONTROL adopted the PREDICT system to forecast the pre-tactical traffic load \cite{liu2018predicting}. The core component of these systems is the aircraft trajectory prediction model that is being reshaped by transformative and disruptive new data-driven techniques especially machine learning in recent decades \cite{shi20204}. 

\begin{figure}
    \centering
    \subfigure[]
    {
        \begin{minipage}{5cm}
            \centering
            \includegraphics[scale=0.45]{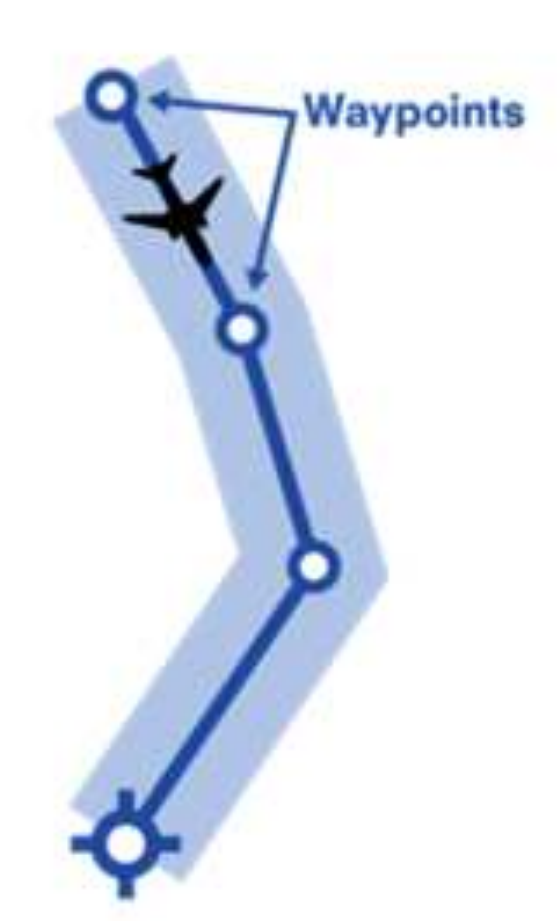}
        \end{minipage}
        \label{fig:RNAV}
    }
    \subfigure[]
    {
        \begin{minipage}{9cm}
            \centering
            \includegraphics[scale=0.45]{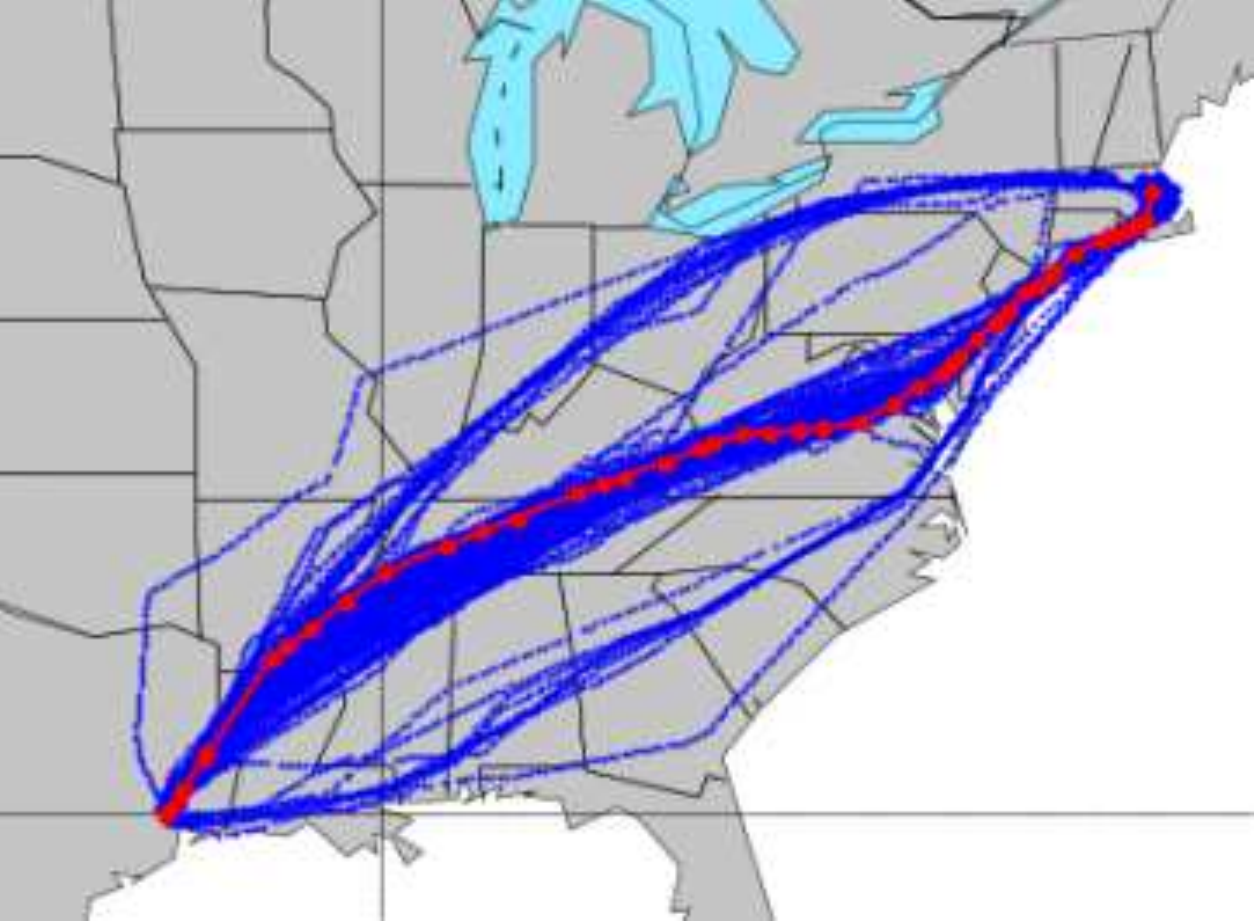}
        \end{minipage}
        \label{fig:devFlight}
    }
    
    \caption{The schematic diagram of air navigation. (a) Area navigation routes follow defined way-points (source: FAA). (b) Actual flight trajectories vs flight plan \cite{liu2018predicting}. Blue lines refer to all trajectories filed the same flight plan shown by the red curve, which reveals that the actual flight trajectories may greatly deviate from their pre-departure plan.}
    \label{fig:airNav}
\end{figure}

In general, an aircraft should obey regulations governing aspects of civil aviation operations and make an airway flight. Airway shown in Figure \ref{fig:RNAV} is a defined corridor consisting of waypoints that connects one specified location to another at a specified altitude, and an aircraft should fly alone the centerline within a specific altitude block and corridor width ideally. Therefore, most of today's practical trajectory prediction systems and literature \cite{liu2011probabilistic, lin2018algorithm, shi2018lstm, lin2019deep, shi20204, barratt2018learning} estimate the flight trajectory in a deterministic way. To be specific, the trajectory is determined based on filed flight plan and historical routes. However, in some cases, the flight status should be adjusted appropriately according to surrounding weather, traffic conditions, and geometry environment \cite{shi20204} as shown in Figure \ref{fig:devFlight}. To address the problem of sensitivity of aircraft to external factors in flight trajectory forecasting, some researchers introduce weather information in modeling \cite{ayhan2016aircraft, liu2018predicting, pang2020conditional}. For example, \cite{liu2018predicting} describes a feature cube referencing system that combines each track point of aircraft and weather conditions like wind speeds and air temperature in front of it to construct a high-dimension feature cube. This feature cube can be treated as a multi-channel image to be abstracted by the Convolutional Neural Network (CNN) and be embedded in to the training process. Although the introduction of weather information makes sense, researches usually employ a single model to estimate the whole trajectory, which conflicts with unique characteristics of flight stages because each flight phase has a definite flight mode and flight requirements \cite{WANG2020103}. Therefore, \cite{shi20204} divides flight trajectories into three phases -- climbing, cruising, and approaching phases using clustering algorithms and proposed a Constrained Long Short-Term Memory (LSTM) network to predict future trajectories in different phases separately in terms of the input sequences. On the contrary, \cite{barratt2018learning} adopted a ruled-based method to separate landing and takeoff phases.

\begin{figure}
    \centering
    \includegraphics[scale=0.08]{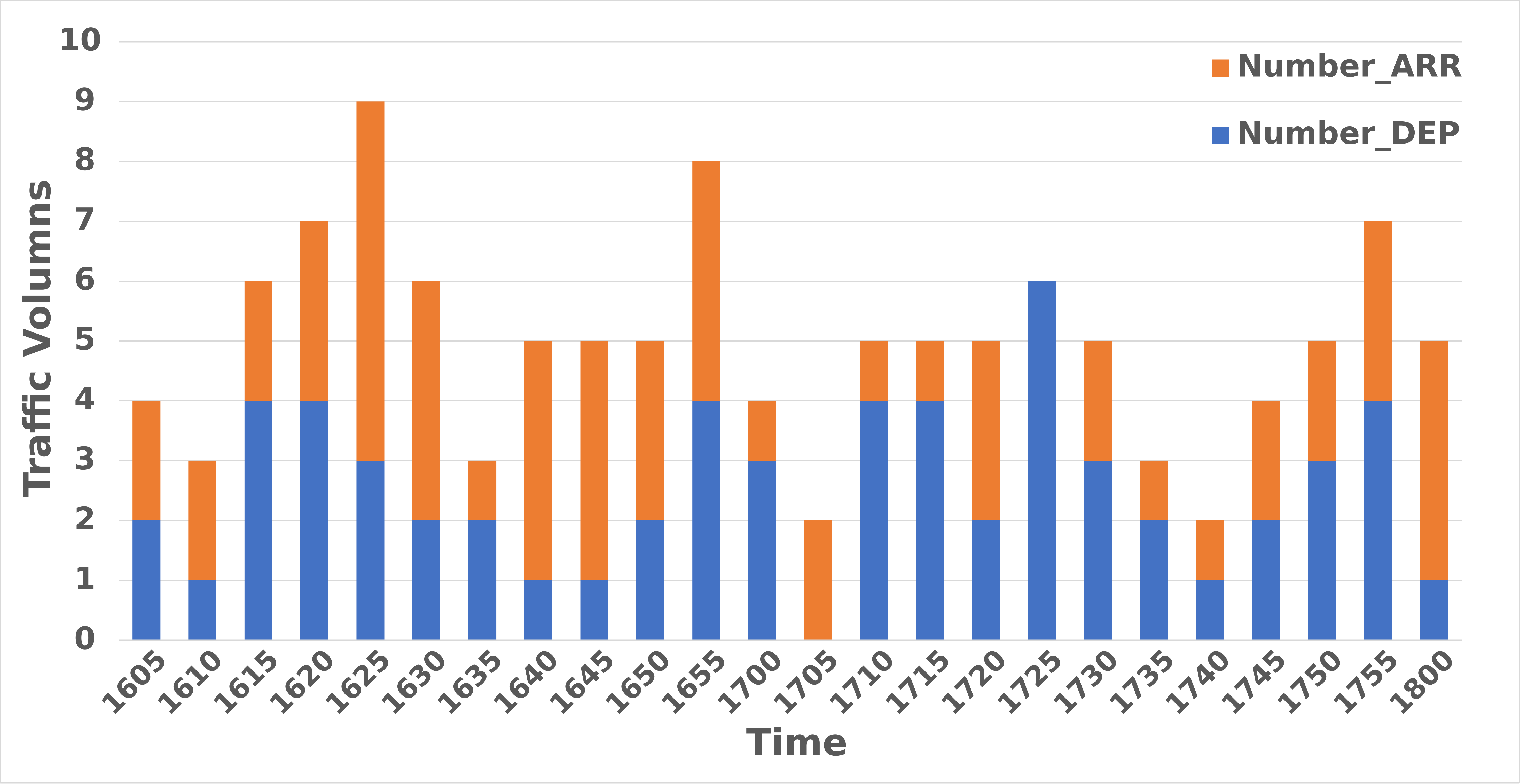}
    \caption{The airline statistics of MCO from 4 pm to 6 pm on December 28, 2019. The orange bar represents the arrival flight while the blue represents the departure flight. (\textit{Data source: Airline On-Time Performance Data from United States Department of Transportation} \cite{AOTP}.)}
    \label{fig:trafficMCO}
\end{figure}

Moreover, pilots have the innate ability to "read" one another. Certain low-altitude airspace is possibly occupied by various private jets, fixed-wing drones and commercial airplanes that take off or land, when they operate the aircraft in this crowded airspace near airports, they obey common sense rules and comply with safety conventions to avoid collisions and it could affect the trajectory. Figure \ref{fig:trafficMCO} is an illustration of the air traffic volumes over a period of time in a day in terms of the number of flights with Orlando International Airport (MCO) -- one of the busiest airports, as origin and destination. We can observe that approximately one airplane take-off or land in a minute, which intuitively presents the congestion of the airspace and indicates the necessity of aircraft-aircraft interaction modeling for flight trajectory prediction in crowded airspace near the airport. 

Previous work barely considers all of above uncertainties comprehensively, hence, in the paper, we propose a phased flight trajectory prediction framework with hybrid RNNs considering dynamic spatio-temporal attributes in real air routes to bridge this gap. Hence in the paper, we present a novel architecture for short to mid-range trajectory prediction in different flight phases, namely takeoff, en-route, and approach. To capture the high-level dynamics of aircraft motion in takeoff/approach phases, we use the spatio-temporal graph (st-graph) to build a set of factor components. Each factor can be represented by a RNN that reflects the structure and interactions of the st-graph. Motion constraints are employed in the trajectory inference process. For en-route trajectory prediction, we take the overall weather features along the planned airway as multiple "channels" to the training process with feature attention and temporal attention, which makes our model adaptable to uncertainty.


To the best of our knowledge, this is the first work proposed to systematically study phased flight trajectory prediction problem, and our work contributes to its advancements in the following ways:

\begin{itemize}
    \item[$-$] We firstly integrate the concept of aircraft-aircraft interaction into flight trajectory prediction framework. The aircraft in crowded airspace are represented by st-graphs, then the use of our developed attention-based structured-LSTM architecture well captures the structure and interactions of st-graph that yields more accurate results for trajectory prediction especially in crowded airspace. 
    \item[$-$] We design a reasonable and effective constraint trigger mechanism that introduces physical aerodynamics limitations in takeoff and approach phases for a more reliable inference process.   
    \item[$-$] We conduct extensive experiments to evaluate our proposed framework on flight trajectory prediction for all phases using our fused dataset. The experimental results demonstrate the performance of our method against some state-of-the-art approaches on our collected and incorporated aviation data. 
\end{itemize}

The remainder of this paper is structured as follows. Section \ref{sec:related_work} highlights the related work. 
Section \ref{sec:overview} formulates the flight trajectory prediction problem and summarizes our work as well as the used datasets. Then the detailed LSTM-based models for takeoff/approach and en-route phases are presented in Section \ref{sec: AAinteraction} and Section \ref{sec:DARNN}, while Section \ref{sec:experiment} evaluates our proposed phased flight trajectory prediction framework using multiple control settings and analyzes the experiment results. Finally, we conclude this paper in Section \ref{sec:conclusion}.




\section{Related Work} \label{sec:related_work}
In this section, we first introduce related studies on mining aviation data, followed by general methods on prediction modeling and trajectory cluster analysis. At last, we summarize past literature in the domain of modeling human-human spatio-temporal interactions. 

\subsection{Aviation Data Mining}
The aviation industry and related fields such as air traffic management systems and weather forecasting generate large-scale complex data, requiring specialized analysis and modeling to extract useful knowledge and insights. In the field of aviation big data analytics and mining, most of researchers focused on operation problems \cite{gallego2018analysis, jung2018data, gariel2011trajectory} and revenue management \cite{warnock2017analysis, an2016map}. In general, the forecasting and analysis of flight delay, flight flow and flight trajectory is the mainstream research for advanced air traffic management.

ADS-B system plays an important role in the safe navigation and air traffic control management, and its messages can be received by everyone when airplanes are flying over. Therefore, ADS-B data has been widely used in the field of aviation data mining. However, considering the lack of security mechanisms in the ADS-B system, researchers applied multiple algorithms to detect anomalies and intrusions of ADS-B messages \cite{leonardi2018ads, habler2018using, tabassum2017ads} based on flight routes or sensors' clocks. 

\subsection{Trajectory Forecasting}
Trajectory forecasting is typically taken as a sequence prediction problem whether it is pedestrian trajectory, vehicle trajectory or flight trajectory. The methods applied in trajectory forecasting are summarized as parametric and non-parametric models. Parametric models are built on prior known parameters including Origin-Destination matrix \cite{wang2019origin}, Kalman filter \cite{ju2019interaction, pathirana2003mobility}, and Markov model \cite{qiao2014self, fang2019citytracker}, while non-parametric models contain support-vector machine \cite{woo2017lane} and neural networks \cite{nikhil2018convolutional}. 

Recently, RNN and its variants such as LSTM \cite{hochreiter1997long} and Gated Recurrent Units \cite{chung2014empirical} have been proven to be effective for sequence prediction tasks such as machine translation and speech recognition. Therefore, they are also widely employed to predict trajectory and obtain the state-of-the-art results, for example, \cite{chandra2019traphic} used a CNN-LSTM hybrid network to predict the near-term trajectories of agents in dense traffic video. \cite{pang2019recurrent} proposed a RNN approach for aircraft trajectory prediction with weather features, which aims to calibrate the flight plan. 

\subsection{Trajectory Clustering}

Aircraft trajectory clustering aims to identify patterns of air traffic by grouping similar trajectories, which in turn contributes to the airspace planning, air traffic flow management, and flight time estimation \mbox{\cite{xuhao2021trajectory}}. K-means and DBSCAN are commonly used clustering algorithms for point-based data including aircraft waypoints. Some clustering methods exist in the literature for flow identification, where the objective is to determine the set of clusters that best fit the operational air traffic flows within a airspace \mbox{\cite{olive2020detection}}. For example, \mbox{\cite{basora2017trajectory}} proposed a en-route flow analysis framework based on HDBSCAN algorithm \mbox{\cite{campello2013density}}, an extented version of DBSCAN, which extracts a simplified hierarchy consisted only of the most significant clusters from the generated complete density-based clustering hierarchy. \mbox{\cite{olive2019trajectory}} focused on clustering flows around airport with DBSCAN and conducted risk assessment in air traffic safety. As for aircraft ground handling, \mbox{\cite{churchill2019clustering}} took DBSCAN as a core in its algorithmic approach to cluster aircraft taxiing trajecotires.

\subsection{Human-human Spatio-Temporal Interactions}

A pioneering work, namely Social Force model \cite{helbing1995social}, used attractive and repulsive forces in terms of relative distances to capture the interactions between agents and achieves good performance. This idea was later followed by researchers in the field of robotics and \cite{trautman2010unfreezing} pointed out that Social Force can not model the complex crowd behavior like cooperation, then a Gaussian process-based way was proposed.

Unlike previous work using hand-crafted proximity relationships, \cite{alahi2016social} introduced a social pooling layer which allows agents share their spatially proximal information with each other. Besides, the interactions that could occur in a more distant future are expected using LSTM. Inspired by this work, \cite{lee2017desire} took into account scene context, and employed a variational autoendoer to obtain a diverse set of hypothetical trajectory samples, which are then ranked by a scoring module based on RNN. \cite{gupta2018social} presented Social GAN by combining recurrent sequence-to-sequence model and generative adversarial networks (GAN). \cite{sadeghian2019sophie} introduced an GAN-based approach with a physical attention and a social attention that incorporate the influence of all agents in the scene as well as the scene context. 

To represent high-level spatio-temporal structures of human motions, \cite{jain2016structural} used spatio-temporal graph \cite{khodayar2018spatio, bai2019spatio} and transformed it into a feedforward mixture of RNNs that can be jointly trained to model dynamics in spatio-temporal tasks. In this paper, we use a similar structure but with attention and constraint modules. 

\section{Preliminaries} \label{sec:overview}
In this section, we present the problem statement for traffic trajectory prediction. The detailed information of datasets and data preprocessing are described next, and the overall structure of our proposed framework is illustrated in Figure \ref{fig:overview}.

\subsection{Problem Statement}

In this paper, we deal with the problem of flight trajectory prediction in three phases -- climbing, cruising, and approaching phases. In general, the problem of flight trajectory prediction can be viewed as a sequence generation task. Let ($x_t^i, y_t^i, z_t^i$) represents the (latitude-longitude-altitude)-coordinate of aircraft $i$ at time-instant $t$ and let $F_t^i$ represents other features such as weather and aircraft types, our problem can be formulated as: 

Given spatial locations \{($x_t^i, y_t^i, z_t^i$)\} and corresponding features {$F_t^i$} for aircraft $i$ from time 1 to $T_{obs}$, predict their future positions for time instants $T_{obs+1}$ to $T_{pred}$. Note that we focus on short to short-term trajectory prediction, the forecast period is typically less than 10 minutes.

\subsection{Data Sources and Preprocessing}

\begin{figure}
    \centering
    \subfigure[]
    {
        \begin{minipage}{0.48\textwidth}
            \centering
            \includegraphics[height = 4cm]{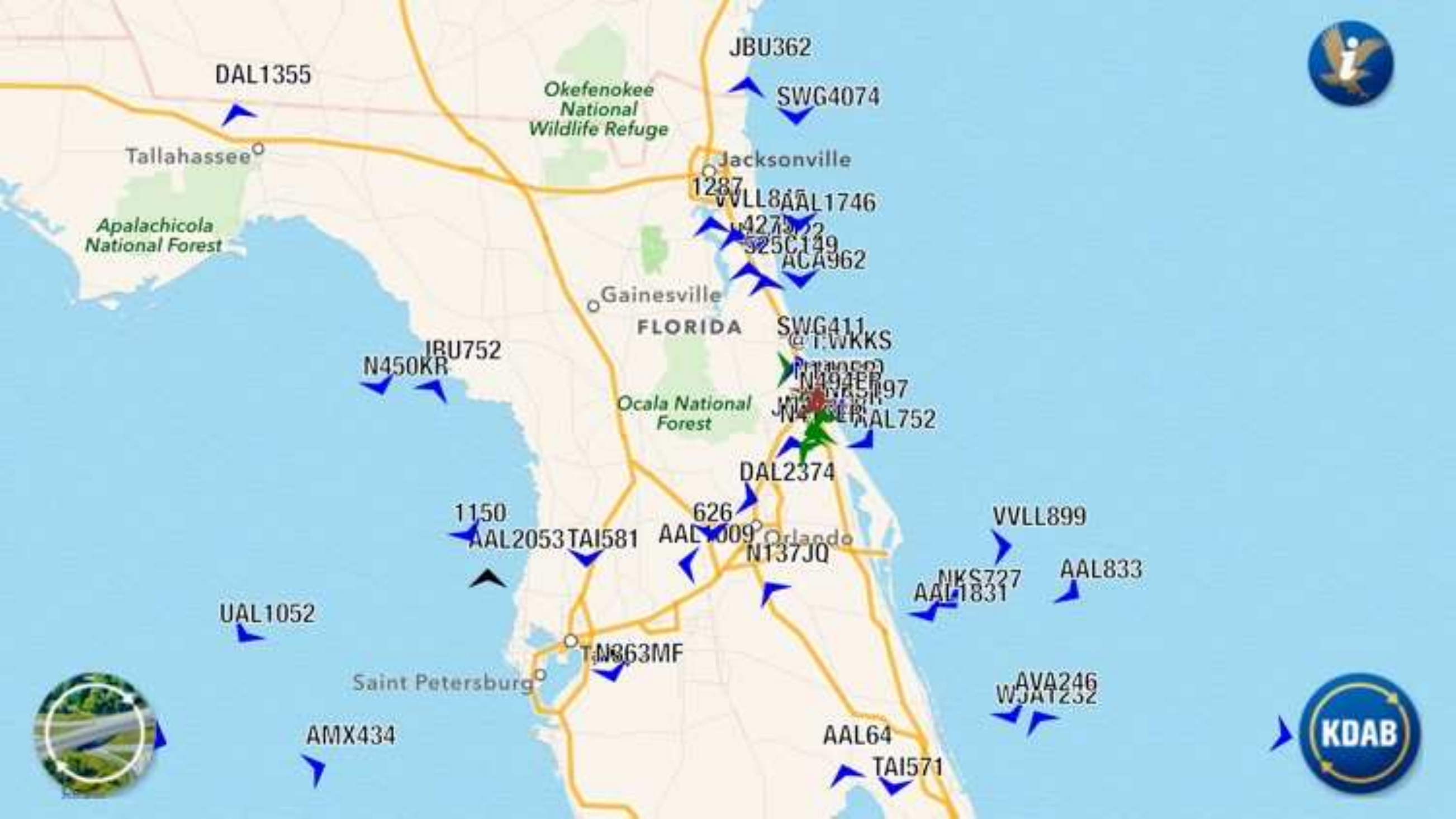}
        \end{minipage}
        \label{fig:erauLive}
    }
    \subfigure[]
    {
        \begin{minipage}{0.48\textwidth}
            \centering
            \includegraphics[height = 4cm]{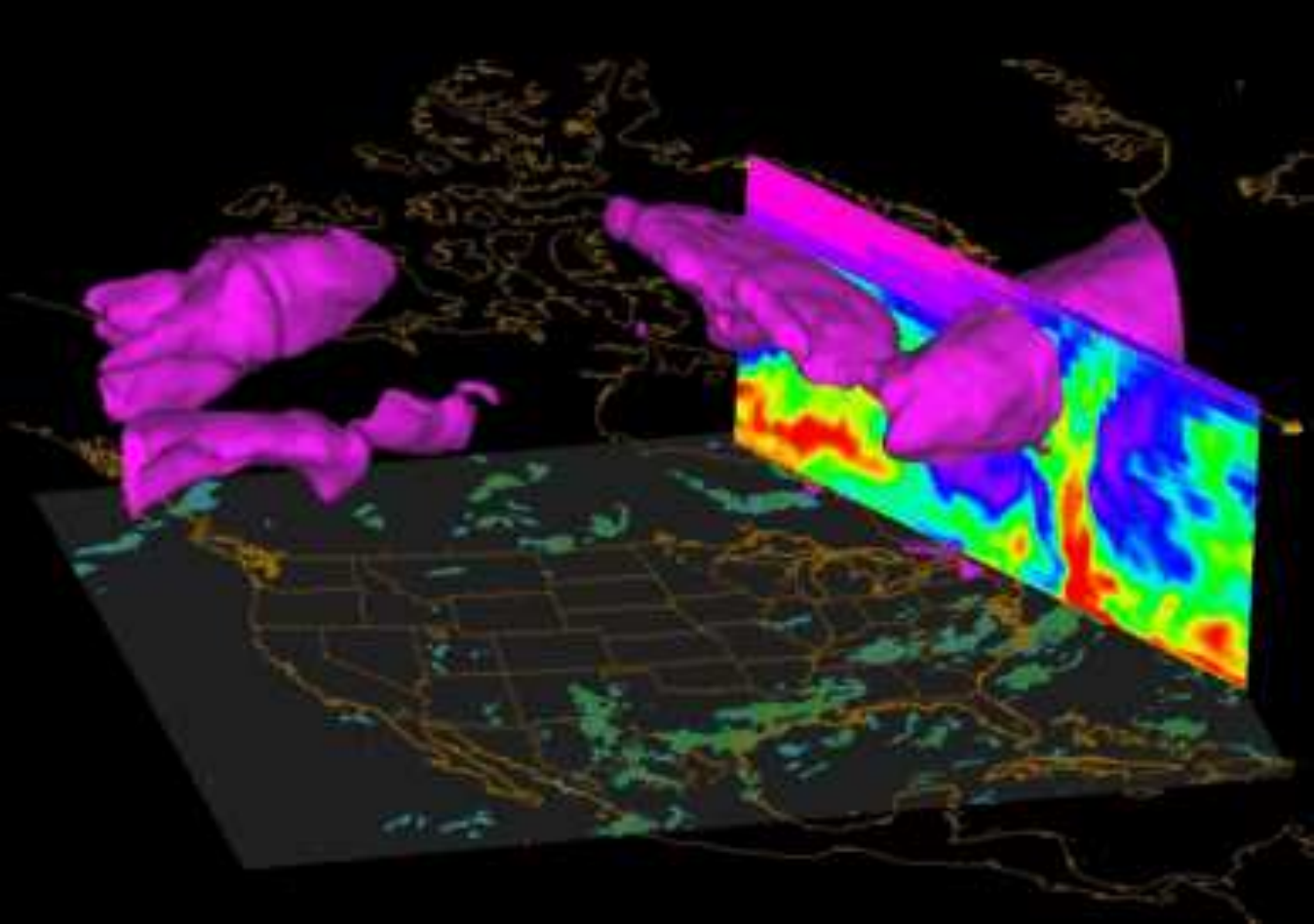}
        \end{minipage}
        \label{fig:rap}
    }
    
    \caption{Data visualization. (a) Airspace condition near Daytona Beach International Airport from the ERAU Live Traffic application developed by ERAU’s NextGen Applied Research Laboratory (NEAR Lab) \cite{ERAUlive}. (b) Vertical profile of relative humidity, red showing high relative humidity and blue showing drier air. (\textit{Data source: the rapid refresh numerical weather model} \cite{RAP}.)}
    \label{fig:dataImage}
\end{figure}

In this paper, we use three types of data -- trajectory data, weather data, and flight profile. The Apache NiFi is utilized to parse and fuse data in a parallel manner. Figure \ref{fig:dataImage} is the graphical representation of some of our data. We will describe their attributes and the corresponding techniques of preprocessing in the following.

\subsubsection{Trajectory data:} ADS–B data is the periodically broadcasted information by the aircraft, ground stations and other aircraft flying in the near airspace to avoid collision. The ADS-B data contains all flights installed with ADS-B equipment which covers most commercial aviation as well as partial general aviation \mbox{\cite{zhang2020spatio}}. ADS-B data used in this paper comes from ADS-B receivers of NEAR Lab at the Daytona Beach campus. In other words, the maximal spatial coverage in the data can only cover the airspace over Florida and some parts of neighboring states. The original data collected by NEAR are ASTERIX Category 033 messages that contain 17 attributes such as latitude/longitude, velocity, aircraft ID and flight number. There are lots of trajectories of trainer aircraft in our data because of many flight schools in Florida. Such trajectories are always unpredictable since student pilots do different actions randomly. Therefore, we filter these trajectories in terms of aircraft ID and aircraft type but keep other private jet's trajectories because they are much more deterministic.

\subsubsection{Weather data:} The weather data used in this paper is High-Resolution Rapid Refresh (HRRR) which is the newest version of the Rapid Refresh (RAP) numerical weather model running by the National Centers for Environmental Prediction (NCEP). HRRR generates data down to a 3-km resolution grid every hour so users can analyze the weather condition for quite small regions of interest. The data is composed of various attributes at different height level. 

\subsubsection{Flight profile:} Flight profile consists of aircraft profile such as aircraft's type and flight plan. Flight plans indicate the planned air route prior to departure. This information is extracted from the records of Traffic Flow Management System (TFMS) system which is also provided by NEAR lab. Original flight plan data is the set of variable-length strings or fixed way-point IDs, so we convert them as pairs of latitude and longitude using the official lookup table \cite{waypoints}. We employ the approximate trajectory partitioning algorithm \mbox{\cite{lee2007trajectory}} to identify characteristic points in flight plans with varying lengths. The final flight plan dataset includes 36 unique flight plans with a fixed length of 10 including the first and last waypoints. It is worth noting that almost all flight plans of general aviation do not exist in the database, we set them as all 0 by default.

\subsubsection{Phase Identification:} There are unique characteristics in each flight phase, however, there is no such a feature in our aerial datasets. Previous work \cite{shi20204} applied clustering algorithms two times to create sub-clusters based on the characteristics of trajectory sequence, but the consistency of generated sub-clusters can not be guaranteed because each entry in the trajectory dataset is relatively close to its neighbors \cite{sun2016large}. To solve this issue, \cite{sun2017flight} proposed to extract continuous flights using DBSCAN 
and then apply fuzzy logic to segment flight trajectory. This method is employed in our work because it can handle large-scale data efficiently. To be specific, we use the following relationships to identify the correct flight pahse given altitude $H$, speed $V$, vertical rate $RoC$:
\begin{subequations}
    \begin{equation}
        \text{if} \; H_{l o} \wedge V_{\text{mid}} \wedge RoC_{+} \; \text{then} \; \textbf{Takeoff}
    \end{equation}
    \begin{equation}
        \text{if} \; H_{h i} \wedge V_{h i} \wedge RoC_{0} \; \text{then} \; \textbf{En-Route}
    \end{equation}
    \begin{equation}
        \text{if} \; H_{l o} \wedge V_{\text{mid}} \wedge (RoC_{-} \vee RoC_{0})  \; \text{then} \; \textbf{Approach}
    \end{equation}
\end{subequations}
where for a commercial flight, $H_{l o} (\eta) = \mathcal{G} (\eta, 10000, 10000)$, $H_{h i} = \mathcal{G} (\eta) (\eta, 35000, 20000)$, $V_{\text{mid}} (v) = \mathcal{G}(v, 300, 100)$, $V_{\text{hi}} (v) = \mathcal{G} (v,600, 100)$, $RoC_{0} (\tau)= \mathcal{G} (\tau, 0, 100)$, $RoC_{+}(\tau) = \mathcal{S} (\tau, 10, 1000)$, $RoC_{-}(\tau)=\mathcal{Z}(\tau, -1000, -10)$. $\mathcal{G}(x;\mu,\sigma)$, $\mathcal{S}(x;a,b)$, $\mathcal{Z}(x;a,b)$ are Gaussian function, S-shaped membership functions, Z-shaped memebership functions, respectively. Fuzzy logic taks such relations between inputs and output to identify the three different flight phase (takeoff, en-route, approach) given data point $(\eta_{i},\tau_{i}, v_{i})$. For more details on phase identification, we refer the reader to \mbox{\cite{sun2017flight}.}

\subsection{Framework Architecture} \label{sec:phase}

\begin{figure}
    \centering
    \includegraphics[scale = .55]{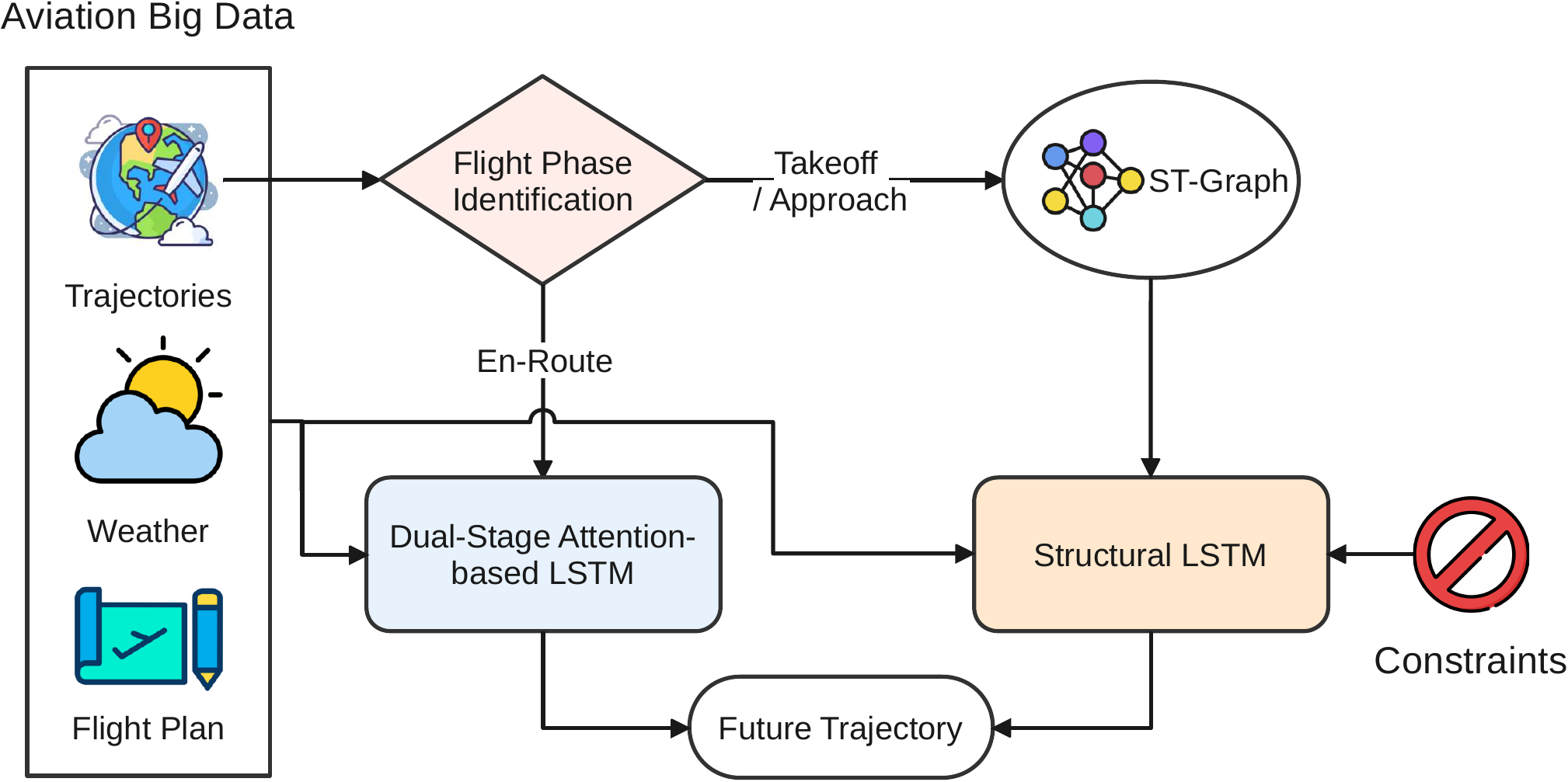}
    \caption{The architecture of our approach.}
    \label{fig:overview}
\end{figure}

Figure \ref{fig:overview} shows the architecture of our approach, which is comprised of two components. First, in the takeoff or approach phase, the aircraft-aircraft interactions are considered to build a spatio-temporal graph where each node represents an aircraft, edges capture the interactions between them. Then, a variant of structural RNN combined with the motion constraints is proposed. We transform the st-graph as a factor graph and represent each factor with a LSTM. The relative importance of each aircraft is captured using a soft-attention module. The second stage is to predict en-route flight trajectory. It is not reasonable to take only local (current position of an aircraft) weather information to predict en-route flight because the air route may change when weather along the flight plan changes. Hence, we leverage CNN to extract compact representation of global (the southeastern region of the United States in this paper) weather information. To appropriately select the most relevant features and capture long-term temporal dependencies of a sequence, the dual-stage attention \cite{qin2017dual} is used. As the last step, the long-term trajectory can be predicted in a multi-step way. 



\section{Modeling Aircraft-aircraft Interactions in crowded airspaces} \label{sec: AAinteraction}
Predicting the motion of aircraft targets while taking into account the complex and often subtle interactions that occur between aircraft in crowded airspaces, is an extremely challenging problem. This problem is inherently comprised of factors in space and time. To capture the dynamics in such a spatio-temporal task, we use factor graph of st-graph representation in which the nodes represent the problem components, and the edges capture their spatio-temporal interactions \cite{jain2016structural}.

\subsection{Spatio-Temporal Graph Representation} \label{sec:st-graph-rep}
Let $\mathcal{G}=\left(\mathcal{V}, \mathcal{E}_{S}, \mathcal{E}_{T}\right)$ denote an st-graph, where $\mathcal{V}$ is the node set, $\mathcal{E}_S$ and $\mathcal{E}_T$ represent as the set of spatial edges and temporal edges, respectively. The graph structure $(\mathcal{V}, \mathcal{E}_{S})$ is unrolled over time through $\mathcal{E}_{T}$ to form $\mathcal{G}$. In the unrolled st-graph, different nodes $v \in \mathcal{V}$ at the same time instant are connected by spatial edges $e = (u, v) \in \mathcal{E}_{S}$, whereas same nodes at adjacent time instants are connected by temporal edges $e = (u, u) \in \mathcal{E}_{T}$. 

\begin{figure}
    \centering
    \includegraphics[scale = .6]{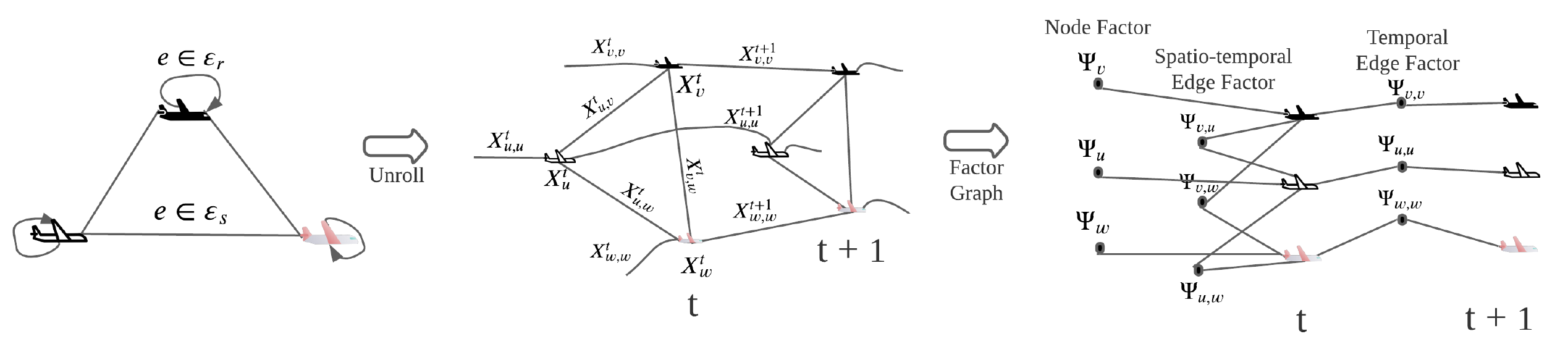}
    \caption{An example st-graph (left), unrolled st-graph (middle), and factor graph (right) of aircraft motions.}
    \label{fig:st-graph}
\end{figure}

In our work, we formulate the problem of flight trajectory prediction in takeoff and descending phases as an st-graph. Hence, the nodes of st-graph represent the aircraft, the spatial edges link two different aircraft at the same time instant, and temporal edges link the identical aircraft at neighbouring time instants. Note that the st-graph is constructed using nodes within a certain spatial scene. The dynamics of relative orientation and distance between aircraft is extracted using spatial edges, while the dynamics of the aircraft's own trajectory is extracted using temporal edges. On the basis of the above definitions, we let \textbf{x}$_v^t$ and \textbf{x}$_e^t$ denote the feature vectors associated with nodes and edges, respectively as shown in Figure \ref{fig:st-graph}, then our goal is to predict the node labels or position vectors in the problem formulation at each time instant $t$. For example, in aircraft-aircraft interaction, the node features can represent their current positions with aircraft attributes and corresponding weather information, and edge features represent their relative orientation; the node labels represent the aircraft future motion that is affected by both its node and edges, leading to a complex system. Such interactions are commonly parameterized with a factor graph that transports how a complex function over st-graph factorizes into a set of simpler functions including a factor function for each node and a pairwise factor function for each edge and the parameters of these factors are needed to be learned \cite{vemula2018social}. In our formulation, we let all the spatial edges share the identical factor function while temporal edges share the other one common factor to reduce the computational complexity and improve the model scalability with more nodes without increasing the number of parameters. 

%

\subsection{Attention-based Structural LSTM Based on Spatio-Temporal Graphs}
\begin{figure}
    \centering
    \includegraphics[scale=0.5]{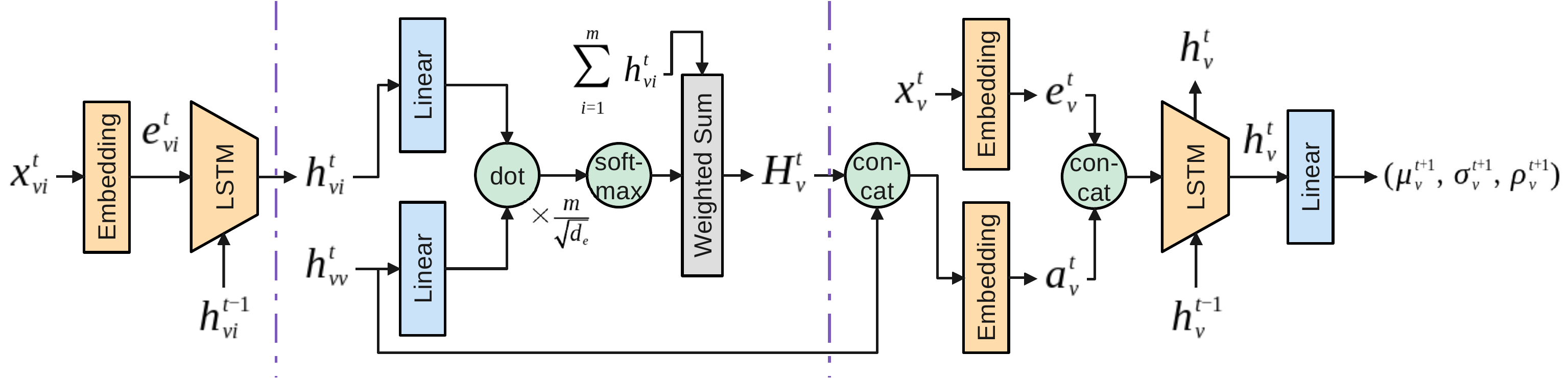}
    \caption{Architecture of Attention-based Structural LSTM that is composed of EdgeLSTM (left), Soft Attention Module (middle), and NodeLSTM (right).}
    \label{fig:S-RNN}
\end{figure}

After the st-graph representation is decomposed into a set of factor components as discussed in Section \ref{sec:st-graph-rep}, we represent each factor using a LSTM similar to \cite{jain2016structural} but a soft attention mechanism is introduced to compute the attention score over hidden states of adjacent spatial edgeLSTMs for each node, which is summarized in Figure \ref{fig:S-RNN}. In detail, we have edgeLSTMs {$\textbf{R}_{\mathcal{E}}$} (the LSTMs related to edge factor $\Psi_{\mathcal{V}}$) to model the dynamics of aircraft-aircraft interations and individual motions in crowded airspace. Then nodeLSTMs {$\textbf{R}_{\mathcal{V}}$} (the LSTMs related to node factor $\Psi_{\mathcal{V}}$) can predict the aircraft's future position at the next time step using the node features and hidden states of the adjacent edgeLSTMs. Besides, the model parameters are shared  across all nodes and edges, therefore, the number of parameters is independent of the number of aircraft at any given time. 


\subsubsection{EdgeLSTM:} For each spatial edgeLSTM {$\textbf{R}_{uv}$} at each time instant $t$, we embed the corresponding edge's feature vector \textbf{x}$_v^t$ into a fixed-length vector $e_{uv}^t$ that is the input to the LSTM cell in the following:
\begin{equation}
    e_{uv}^t = \phi ({\rm x}_{uv}^t; W_{Spatial}^e)
\end{equation}
\begin{equation}
    h_{uv}^t = {\rm LSTM} (h_{uv}^{t-1}, e_{uv}^t; W_{Spatial}^r)
\end{equation}
where $\phi(\cdot)$ is an embedding function, $W_{Spatial}^e$ is the embedding weight. The edgeLSTM weights are denoted by $W_{Spatial}^r$ and the hidden state is denoted by $h_{uv}$. Similarly, the LSTM cell for temporal edgeLSTM {$\textbf{R}_{uu}$} is built as:
\begin{equation}
    e_{uu}^t = \phi ({\rm x}_{uu}^t; W_{Temporal}^e)
\end{equation}
\begin{equation}
    h_{uu}^t = {\rm LSTM} (h_{uu}^{t-1}, e_{uu}^t; W_{Temporal}^r)
\end{equation}
these weights are the trainable parameters for edgeLSTMs.

\subsubsection{Soft Attention Module:} \label{sec:attention} We compute a scaled dot-product attention  \cite{vaswani2017attention} given in Equation (\ref{eq:attention}) over the hidden states of edgeLSTMs of all spatial edges $h_{vi}^t, i \in [1, m]$ which each node associates with. 

\begin{equation} \label{eq:attention}
    {\rm Attention} (h_{vv}^t, h_{vi}^t) = \sigma (\frac{m}{\sqrt{d_e}}\left\langle W_{vv} h_{v v}^{t}, W_{v} h_{vi}^{t}\right\rangle) h_{vi}^t 
\end{equation}
where $m$ is the number of spatial edges the node connects,  $W_{vv}$ and $W_{v}$ are weights that are used to project the hidden states into vectors with a fixed-dimension $d_e$. $\sigma$ is the softmax function. Note that $m$ depends on the number of aircraft and may change in every time frame. 

Therefore, we can obtain the output vector of the attention module, a weighted sum of hidden states at each time step $t$,
\begin{equation}
    H_v^t = \sum_{i=1}^{m} {\rm Attention} (h_{vv}^t, h_{vi}^t)
\end{equation}

\subsubsection{NodeLSTM:} The aim of nodeLSTMS is the position estimation, to be specific, is to predict the distribution of the flight trajectory $(\hat{x}, \hat{y}, \hat{z})$ at the next time step based on concatenated information at the previous time frame. Inspired by the previous work \cite{graves2013generating, alahi2016social}, we assume a multivariate Gaussian distribution and we denote $(\mu_v^t, \sigma_v^t, \rho_v^t)$ as the mean, standard deviation, and correlation coefficient of it. These parameters are predicted by a linear layer $W_p$ and the predicted coordinates $(\hat{x}, \hat{y}, \hat{z})_v^t$ are given by
\begin{equation}
    (\hat{x}, \hat{y}, \hat{z})_v^t \sim \mathcal{N} (\mu_v^t, \sigma_v^t, \rho_v^t)
\end{equation}
\begin{equation}
    \mu_v^t = (\mu_x, \mu_y, \mu_z)_v^t, \ \sigma_v^t = (\sigma_x, \sigma_y, \sigma_z)_v^t, \ \rho_v^t = (\rho_x, \rho_y, \rho_z)_v^t
\end{equation}
\begin{equation}
    [\mu_v^t, \sigma_v^t, \rho_v^t] = W_p h_v^t
\end{equation}

The input to nodeLSTM consists of the embedded node's features ${\rm x}_v^t$, the hidden state of the corresponding temporal edgeLSTM $h_{vv}^t$, and the attention output $H_v^t$ as follows:
\begin{equation}
    e_v^t = \phi({\rm x}_v^t; W_{node}^e)
\end{equation}
\begin{equation}
    a_v^t = \phi ({\rm concat}(h_{vv}^t, H_v^t); W_{attention}^e)
\end{equation}
\begin{equation}
    h_v^t = {\rm LSTM} (h_v^{t-1}, {\rm concat}(e_v^t, a_v^t); W_{node}^r)
\end{equation}
The trainable parameters of the LSTM  model are jointly learned by minimizing the negative log-likelihood loss ($L_v$ for the trajectory of node $v$):

\begin{equation}
\begin{aligned}
    L_v (W_{Spatial}^e, W_{Spatial}^r, W_{Temporal}^e, W_{Temporal}^r, W_{node}^e, W_{attention}^e, W_{node}^r, W_p) \\
    = - \sum_{t = T_{obs}+1}^{T_{pred}} {\log (\mathbb{P}(x_v^t, y_v^t, z_v^t | \mu_v^t, \sigma_v^t, \rho_v^t))}
\end{aligned}
\end{equation}

\subsection{Motion Constraints and Model Inference}
During test time, the trained model is fitted to sample the predicted multivariate Gaussian distribution to obtain the future position \{$(\hat{x}_v^t, \hat{y}_v^t, \hat{z}_v^t)$\} using the observation at previous time steps from $t$ to $T_{obs}$. From time step $T_{obs+1}$ to $T_{pred}$, the predicted positions \{$(\hat{x}_v^t, \hat{y}_v^t, \hat{z}_v^t)$\} are used to compute node features and spatial edge features in place of actual positions for each time. 

To mitigate the generalization error especially the effect of noise and variance of the model, we propose knowledge-guided motion constraints for takeoff and approach phases independently and incorporate them into the inference process as shown in Algorithm \ref{alg:constraint}. The sets of parameters $\alpha$ and $\beta$ in the formulations of motion constraints are achieved and explained in Section \ref{sec:constriant1} and \ref{sec:constriant2}. In addition, the phase identification has been discussed in Section \ref{sec:phase}. Note that some information such as heading may be missing in the trajectory data so that we can not calculate the constraint parameters. Therefore, the motion constraints module will not be applied in such a case.

\begin{algorithm} 
\caption{Flight Trajectory Inference Based on Constraints}
\begin{algorithmic}[1] \label{alg:constraint}
\renewcommand{\algorithmicrequire}{\textbf{Input:}}
\renewcommand{\algorithmicensure}{\textbf{Output:}}
\REQUIRE Trajectory $\{(x_t, y_t, z_t)\}$ with associated features $F_t$, $t = 1,...,T_{obs}$, a set of constraints $\alpha, \beta$ for takeoff and approach phases, respectively.
\ENSURE Future trajectory $\{(x_t, y_t, z_t)\}$, $t = T_{obs+1},...,T_{pred}$
\STATE $\mu_v^t, \sigma_v^t, \rho_v^t \leftarrow$ fit the trained LSTM model with test input ${(x_t, y_t, z_t, F_t)}, t \in [1, T_{obs}]$
\STATE $phase \leftarrow$ phase indicator function with test input
\\ \textit{LOOP Process}
\FOR {$t = T_{obs+1}$ to $T_{pred}$}
\WHILE {$phase$ = takeoff or approach}
\STATE $(x_{t+1}, y_{t+1}, z_{t+1}) \leftarrow$ sample from the predicted multivariate Gaussian distribution $(\mu_v^t, \sigma_v^t, \rho_v^t)$ 
    \IF {$(x_{t+1}, y_{t+1}, z_{t+1})$ satisfies limitation $\alpha$ or $\beta$}
    \STATE Append $(x_{t+1}, y_{t+1}, z_{t+1})$ to the estimated future trajectory sequence $\{(x, y, z)\}$
    \STATE $t = t+1$
    \\ \textit{break}
    \ELSE \STATE \textit{continue}
    \ENDIF 
\ENDWHILE
\ENDFOR
\RETURN {estimated trajectory $\{(x,y,z)\}$}
\end{algorithmic}
\end{algorithm}
 
\subsubsection{Climbing and Descending Constraints}\label{sec:constriant1}

\begin{figure}
    \centering
    \includegraphics[scale = 0.55]{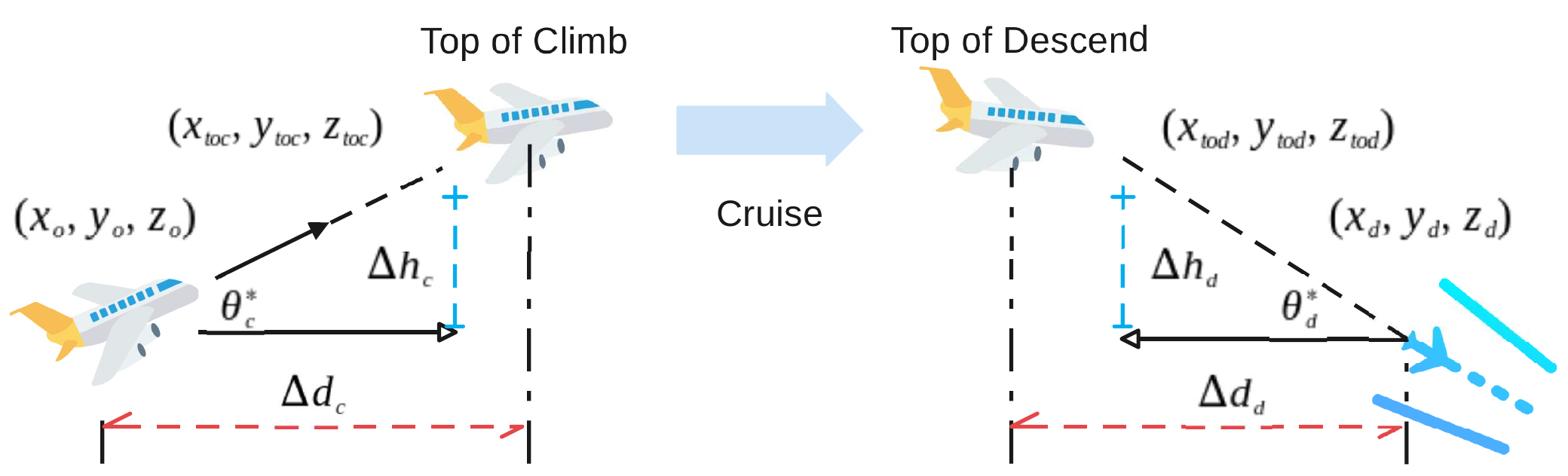}
    \caption{Graphical illustration of the Climbing/Descending Constraint Formulation.}
    \label{fig:constriant}
\end{figure}

Figure \ref{fig:constriant} depicts a schematic diagram of climbing and descending phases. The constraints are considered in the paper to apply the angle limitation. We firstly extract all trajectories that are identified as in the takeoff phase from the entire data, and set the points with maximum altitudes as their Top of Climb (TOC) points $\{(x_{toc}, y_{toc}, z_{toc})\}$ to calculate the climb angle $\{\theta_{c}^*\}$ based on the first way-points $\{ (x_{o}, y_{o}, z_{o}) \}$ above 1500 feet of the trajectory. Then the maximum climb angle $\theta_c = \max \{\theta_c^*\}$ can be known using Equation (\ref{eq:cd_constraints}). Similarly, we can also obtain the Top of Descend (TOD) points $\{(x_{tod}, y_{tod}, z_{tod})\}$ and the maximum descend angle $\theta_d$. 
\begin{equation}
\left\{
             \begin{array}{lr}
             \theta \leq \theta_c = \max \limits_{\theta_c^*} \{ \theta_c^* = \tan^{-1} (\Delta h / \Delta d)  \}, \\
             \Delta h = |z_{top} - z_{o}|, \\
             \Delta d=2 r \sin ^{-1}(\sqrt{\sin ^{2}\left(\Delta y / 2\right)+\cos \left(y' \right) \sin ^{2}\left( \Delta x / 2\right)}), \\
             \Delta x =  |x_{top} - x|, \Delta y =  |y_{top} - y|,\\
             \cos \left(y' \right)=\cos \left(y_{top} \right) \cos \left(y \right)
             \end{array}
\right.
\label{eq:cd_constraints}
\end{equation}
\noindent where $\Delta d$ is the great-circle distance between two way-points on a sphere with latitude and longitude ($x$, $y$) and $r$ is the radius of the Earth.

\subsubsection{Rate of Turn Constraint}\label{sec:constriant2}
The rate of turn (ROT) is the number of degrees (expressed in degrees per second) of heading change that an aircraft makes, the ROT can be determined by taking the constant of 1,091,
multiplying it by the tangent of any bank angle $\theta_b$ and dividing that product by a given airspeed $v_a$ in knots \cite{federal2009pilot}. Suppose the maximum ROT is $\omega_{ROT}$, then the ROT constraint can be expressed as Equation (\ref{eq:ROT}) that also describes the relationship between bank angle and airspeed as they affect the ROT. 
\begin{equation}
\left\{
             \begin{array}{lr}
             \omega \leq \omega_{ROT} = \max  \{ \omega_{ROT}^*  \}, \\
             \omega_{ROT}^* = (1091 \times \tan{\theta_b}) / v_a
             \end{array}
\right.
\label{eq:ROT}
\end{equation}
where $\{ \omega_{ROT}^*  \}$ can be obtained from our data in the same way as described in Section \ref{sec:constriant1}.

\section{Dual-stage Attention-based LSTM for En-route Trajectory Prediction} \label{sec:DARNN}
En-route is a flight phase between climbing and descending which usually consumes the majority of a flight. Typically, aircraft in the en-route phases fly at a constant velocity and altitude but their heading or direction of flight may change. Due to the vast airspace and flight level allocation by ATC in the en-route phase, an airplane has relatively little impact on others. Intuitively, it seems much easier to predict flight trajectory in en-route than other phases. However, because of the requirement of global observations instead of local observations of weather uncertainty, it is a big challenge to predict the abnormal but actual trajectory that deviates from original flight plan.  Hence, the dual-stage attention mechanism \cite{qin2017dual}, in which the relevent input features and relevant hidden states are adaptively and sequentially extracted, can be deployed to our en-route trajectory prediction model. For example, each feature vector in the first attention layer will be weighted so that more information of more important features (higher weights) can be introduced into the decoder. Then, in the second stage, important time steps such as the one that exists a sudden thunderstorm are captured by the temporal attention layer.

\subsection{Weather Feature Representation} \label{sec:weatherCube}
The weather information in the Rapid Refresh model is represented by a set of reference 3D points including latitude, longitude, and altitude. The latitude and longitude of reference points are fixed. Therefore, we build the boundary in the latitude-longitude plane using the reference points and the middle points between each neighboring reference point, and construct a series of cubes in terms of altitude. A cube contains the weather information of a 3D airspace such as humidity, wind speed and direction. We assume that an aircraft flies at a fixed altitude in the en-route phase, in other words, this aircraft passes through cubes on a horizontal plane. Therefore, to extract the high-dimensional global weather information, each feature can be thought as a channel and be filtered into a fixed-size compact representation using three 2-D convolutional layers with dense. There are 7 channels or weather features in total, they are HGT (geopotential height), TMP (temperature), RH (relative humidity), VVEL (vertical velocity), UGRD (u-component of wind), VGRD (v-component of wind), and ABSV (absolute vorticity). The output at time step $t$ is denoted by:
\begin{equation}
    C_t = {\rm conv} (F_{W}^t; W_{cnn})
\end{equation}
where $F_{W}^t$ represents weather information of all channels, $W_{cnn}$ represents trainable parameters of convolutional neural networks. Note that the dimension of weather feature representation should be equal to the input sequence.

    

\begin{figure}
    \centering
    \includegraphics[scale=.5]{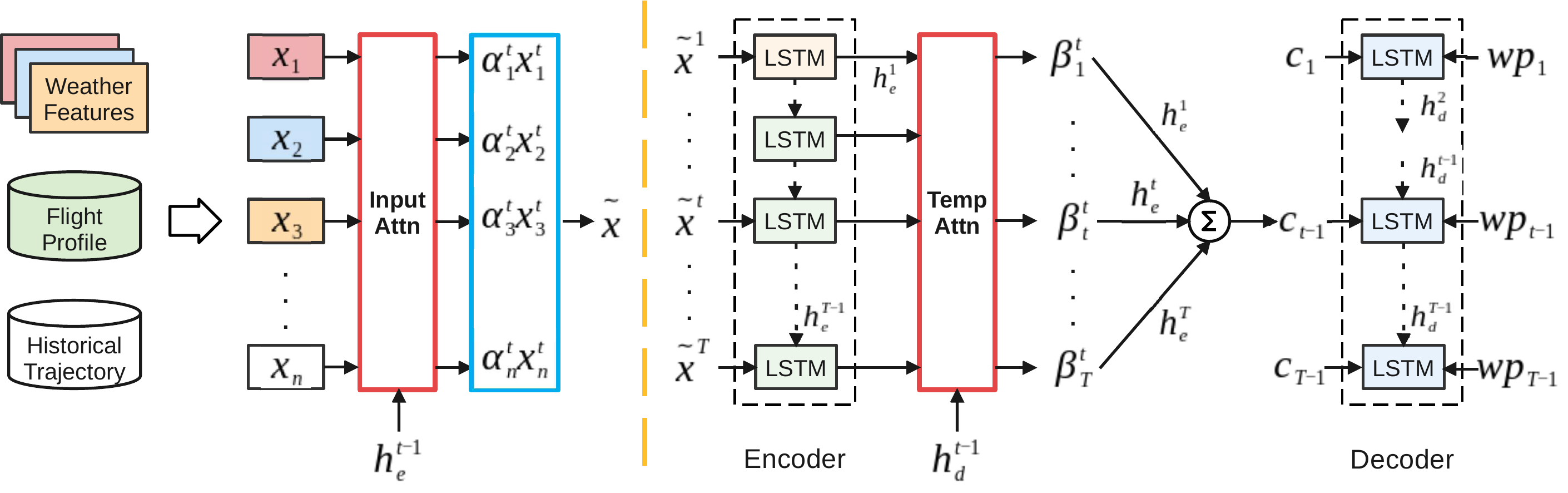}
    \caption{Overview of En-route trajectory prediction. At time-step $t$, input attention module (left) computes attention weights $\{\alpha\}$ for each input feature $\{x_1, x_2, ..., x_n\}$ according to the hidden state $h_e$ of the encoder at previous time-step $t-1$. The weighted input is then fed into the encoder. Temporal attention module (right) computes attention weights $\{\beta\}$ according to the hidden state $h_d^{t-1}$ of the decoder. The weighted sum of encoder hidden states over all time-steps $c_{t}$ is represented as the input to decoder that outputs predicted way point $\hat{wp}_T$}
    \label{fig:fig:dualAttn}
\end{figure}

\subsection{Dual-stage Attention Mechanism}

Figure \ref{fig:fig:dualAttn} shows the feedward-pass of our proposed en-route trajectory prediction model based on dual-stage attention-based LSTM. The input attention and temporal attention blocks have the same score function as we used in Section \ref{sec:attention}.

Different from the structrual LSTM for low-altitude airspace, normal en-route trajectory usually follows the flight plan without influence caused by other aviation, consequently, we need to focus more on the aerodynamics of flight. We thus take the velocity components in each direction $\vec{V}$ and the track angle $\Theta$ into account to design the Location-Velocity-Angle (LVA) loss function given by:
\begin{equation}
    L(W_{cnn}, W_{rnn}, W_e) = \frac{1}{n} \sum_{i=1}^{n} [(\vec{V}_i - \hat{\vec{V}}_i)^2 + (\Theta_i - \hat{\Theta}_i)^2 + ({\rm T}_i - \hat{{\rm T}}_i)^2]
\end{equation}
where $W_{rnn}$ refers to the parameters of LSTMs (in both encoder and decoder) to learn,  $W_e$ is the weight of embedding layer that is used to get a fixed-dimension representation of the flight profile. ${\rm T}_i - \hat{{\rm T}}_i$ denote the distance between two trajectories, which can be calculated by the Equation (\ref{eq:cd_constraints}).

\section{Experiments} \label{sec:experiment}
In this section, we firstly describe the experimental settings, and then compare the performance of our proposed framework with other methods. Next, we will use two example to explain why our method can achieve good performance.

\subsection{Evaluation Metrics}
We use the aerial data within the southeastern region around Florida state of the United States in 2019 to assess the performance of our proposed framework. The data is segmented into three sub-datasets for different flight phases. For each sub-dataset, the training set and test set are partitioned according to specific dates which are sampled randomly. Specifically, the data of 219 days (\%60 of 365) is taken as the training set, the data of 73 days (\%20 of 365) is taken as the validation set, while the remaining data of other days is taken as the test set. We also use the same training and testing dataset for other baseline and state-of-the-art methods for performance comparison. 
Three commonly-used evaluation metrics are applied to report the prediction error:
\begin{itemize}
    \item \textbf{Average displacement error (\textbf{ADE}):} the mean euclidean distance (in meter) over all predicted way-points and actual trajectory.
    \item \textbf{Final displacement error (\textbf{FDE}):} the distance (in meter) between final way-point of the estimated flight trajectory and the actual destination.  
    \item \textbf{Mean absolute error (\textbf{MAE}:)} in this paper, this metrics is mainly employed to measure the prediction error of a single parameter of geo-locations. The latitude and longitude values are decimal degrees and altitude is in meters.
\end{itemize}

\subsection{Implementation Details}

The dimension of hidden states of all LSTMs in our framework is set to 256, while the dimension of embedding layers in structrual LSTM is set to 64 with ReLU activation and dimension of embedding layers in en-route depends on input sequence. 
The ADS-B system records data per 0.5 second so that the distance between two adjecent way-points is too close. Hence, we scale the sampling rate to let one time-step corresponds to 10 seconds. During test time, we observe a trajectory for 3 minutes ($T_{obs} = 18$) and predict is future trajectory for the next 5 minutes ($T_{pred} = 48$). We use a learning rate of 0.005 with Adam optimizer to train our framework on a single GPU with Pytorch. 

\subsection{Results and Analysis}
Because of the limitation of our dataset, there are not too many complete trajectories (from origin to destination). Therefore, the results of our experiments mainly show the medium-term or short-term prediction error, but the longer trajectory forecasting can also be obtained by the mean of multi-step forecasting and the jointed outputs from each model. 
We compare the performance of our model with off-the-shelf methods as well as multiple control settings:

    

\begin{itemize}
    \item Kalman Filter (KF) \mbox{\cite{alahi2016social}}: we extrapolate trajectories with assumption of linear accelaration in takeoff/approach phases and constant speed in en-route phase. KF is applied in the case where the input only contains waypoints (\textbf{C1}).
    \item Hidden Markov Model (HMM) \mbox{\cite{ayhan2016aircraft}}: HMM is applied in both cases where the input only contains waypoints (\textbf{C1}) and the input contains all information including flight plans and weather (\textbf{C2}).
    \item Our Vanilla LSTM (V-LSTM): this is simplified setting of our model where we remove the st-graph, motion constraints, and dual-attention modules and treat all the trajectories to be independent of each other. This model is applied in both cases (\textbf{C1} and \textbf{C2}).
    \item Social LSTM (S-LSTM) \mbox{\cite{alahi2016social}}: it is a state-of-the-art method named Social LSTM (S-LSTM), which takes into account influence of interaction. This model is applied in both cases (\textbf{C1} and \textbf{C2}).
\end{itemize}

\begin{table}[htbp]
\caption{Quantitative Results (C1) of All the Methods for Takeoff Phase}
\begin{tabular}{lccccc}
\toprule
\multirow{2}{*}{Model} & \multirow{2}{*}{ADE} & \multirow{2}{*}{FDE} & \multicolumn{3}{c}{MAE}                              \\ \cmidrule{4-6} 
                       &                      &                      & longitude       & latitude        & altitude         \\ \midrule
V-LSTM                 & 1842.9849             & 3465.8422            & 0.0188          & 0.0075           & 195.2549         \\
KF                     & 2375.4481            & 4331.8186            & 0.0214          & 0.0093          & 231.7216         \\
HMM                    & 1981.8992            & 2994.1124            & 0.0186          & 0.0074          & 182.4849         \\
S-LSTM                 & 688.6658             & 1114.6335             & 0.0056          & 0.0048         & 147.5498          \\
Our                    & \textbf{357.4578}    & \textbf{512.4812}    & \textbf{0.0024} & \textbf{0.0022} & \textbf{68.2185} \\ \bottomrule
\end{tabular}
\label{tab:takeoff_c1}
\end{table}

\begin{table}[htbp]
\caption{Quantitative Results (C1) of All the Methods for Approach Phase}
\begin{tabular}{lccccc}
\toprule
\multirow{2}{*}{Model} & \multirow{2}{*}{ADE} & \multirow{2}{*}{FDE} & \multicolumn{3}{c}{MAE}                              \\ \cmidrule{4-6} 
                       &                      &                      & longitude       & latitude        & altitude         \\ \midrule
V-LSTM                 & 2155.1877          & 4311.8799            &  0.0199         & 0.0184          & 222.8441          \\
KF                     & 4572.4678            & 6248.5177            & 0.0374          & 0.0266          & 219.5478         \\
HMM                    & 2949.7379            & 4712.5413            & 0.0214          & 0.0193          & 194.5148         \\
S-LSTM                 &  1436.8421            & 3248.2154            & 0.0083          & 0.0057          & 243.5582          \\
Our                    & \textbf{725.7452}    & \textbf{2811.9688}    & \textbf{0.0053} & \textbf{0.0034} & \textbf{151.7838} \\ \bottomrule
\end{tabular}
\label{tab:approach_c1}
\end{table}

\subsubsection{Comparison between Models for Takeoff and Approach Phases}
Table ~{\ref{tab:takeoff_c1}} and Table ~{\ref{tab:approach_c1}} present the prediction errors of climbing and descending flights for all methods in Case 1 during test time, respectively. The models under comparison includes baseline methods without considering the aircraft-aircraft interations such as V-LSTM, KF, and HMM. Besides, S-LSTM, which takes into account influence of interaction, is also used for comparison. In general, our model far exceeds others, especially in the approach phase where there are much more complex motion patterns because of the air traffic control. In Case 2, as shown in Table ~{\ref{tab:takeoff_c2}} and Table ~{\ref{tab:approach_c2}}, we obtain the similar conclusion. Besides, we can demonstrate that more information leads to better predictive trajectories when we comapre results of different phases in a same case (e.g. Table ~{\ref{tab:takeoff_c1}} vs. ~{\ref{tab:takeoff_c2}} and Table ~{\ref{tab:approach_c1}} vs. Table ~{\ref{tab:approach_c2}}).

\begin{table}[]
\caption{Quantitative Results (C2) of All the Methods for Takeoff Phase}
\begin{tabular}{lccccc}
\toprule
\multirow{2}{*}{Model} & \multirow{2}{*}{ADE} & \multirow{2}{*}{FDE} & \multicolumn{3}{c}{MAE}                              \\ \cmidrule{4-6} 
                       &                      &                      & longitude       & latitude        & altitude         \\ \midrule
V-LSTM                 & 573.3794             & 1643.6625            & 0.0042          & 0.0035          & 116.2396         \\
HMM                    & 612.4712            & 1385.1277            & 0.0044          & 0.0038          & 130.8003         \\
S-LSTM                 & 488.9801             & 914.8549             & 0.0036          & 0.0029          & 85.7734          \\
Our                    & \textbf{341.8263}    & \textbf{437.8141}    & \textbf{0.0024} & \textbf{0.0022} & \textbf{60.7959} \\ \bottomrule
\end{tabular}
\label{tab:takeoff_c2}
\end{table}

\begin{table}[]
\caption{Quantitative Results (C2) of All the Methods for Approach Phase}
\begin{tabular}{lccccc}
\toprule
\multirow{2}{*}{Model} & \multirow{2}{*}{ADE} & \multirow{2}{*}{FDE} & \multicolumn{3}{c}{MAE}                              \\ \cmidrule{4-6} 
                       &                      &                      & longitude       & latitude        & altitude         \\ \midrule
V-LSTM                 & 1048.5811          & 2183.5482            &0.0085           &0.0054           & 234.2214         \\
HMM                    & 862.1238            & 1284.5943           & 0.0073          & 0.0046 & 193.4833         \\
S-LSTM                 & 801.6282             & 1422.5314            & 0.0062          & 0.0046          & 158.7524          \\
Our                    & \textbf{450.9431}    & \textbf{766.2355}    & \textbf{0.0036} & \textbf{0.0024} & \textbf{92.8776} \\ \bottomrule
\end{tabular}
\label{tab:approach_c2}
\end{table}

\begin{figure}
    \centering
    \subfigure[]
    {
        \begin{minipage}{0.48\textwidth}
            \centering
            \includegraphics[scale=.6]{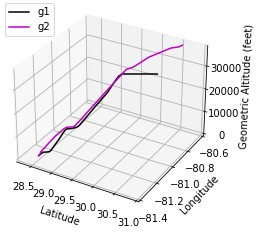}
        \end{minipage}
      \label{fig:flight_overview}
    }
    \subfigure[]
    {
      \begin{minipage}{0.48\textwidth}
            \centering
            \includegraphics[scale=.6]{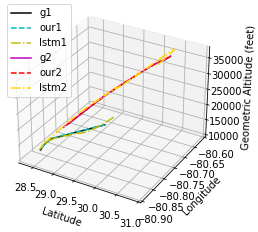}
        \end{minipage}
        \label{fig:d1}
    }
    
    \subfigure[]
    {
        \begin{minipage}{0.48\textwidth}
            \centering
            \includegraphics[scale=.6]{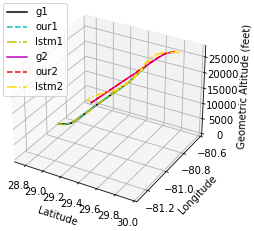}
        \end{minipage}
        \label{fig:d2}
    }
    \subfigure[]
    {
        \begin{minipage}{0.48\textwidth}
            \centering
            \includegraphics[scale=0.6]{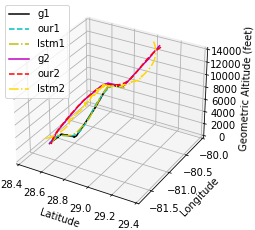}
        \end{minipage}
        \label{fig:d3}
    }    
    \caption{Illustration of our proposed framework predicting trajectories of two aircraft descend simultaneously. (a) Overview of descending trajectories. (b) Descend stage 1. One aircraft $g1$ as depicted in black yields the other one $g2$ which is marked in blue line. (c) Descend stage 2. Two aircraft descend almost together, but the distance between them is increased gradually for collision avoidance. (d) The last descend stage. It is obvious that $g1$ fly at a higher altitude than $g2$ and prepares for descending until $g2$ lands.} 
    \label{fig:croweded_analytics}
\end{figure}

To further explain how the air traffic control and aircraft operations influence flight trajectory as well as what patterns corresponding to ATM learned by models, we then take a case where two airplanes (UPS1328 and FDX1074) approach MCO airport on August 1, 2019 as an example. Figure \ref{fig:croweded_analytics} shows this classical ATM case in a crowded airspace. The process is described in detail in its caption. Besides, we applied our model and vanilla LSTM to predict these two trajectories, separately. The result shows that our model learns how multiple agents cooperate but vanilla LSTM fails. Figure \ref{fig:flight_overview} highlights three Points of Interests (POIs) $\mathcal{I}_1, \mathcal{I}_2$ and $\mathcal{I}_3$, in other words, the way-points that one aircraft yield the other one to ensure the flight safety. From Figure \ref{fig:d1}, we observe that the the trajectory predicted by the Vanilla LSTM $lstm2$ exceeds the final point or destination $\mathcal{D}_1$ in this time frame, and reaches the other aircraft. The similar result can also be found in Figure \ref{fig:d3}. These phenomenons demonstrate that the interaction modeling makes sense in the flight trajectory prediction problem in low-altitude and crowded airspace.

\subsubsection{Comparison between Models for En-route Phase}
Table ~{\ref{tab:en-route_c1}} shows the performance of each model in En-route phase in Case 1 during test time. Generally speaking, the prediction errors of en-route phase are much less than the prediction errors of other two phases because of the relative stationarity of trajectories in en-route phases in practice. HMM models can predict the trajectory more accurately than others while our model performs badly to forecast longitude and latitude. This phenomenon may be the overfitting because of the high model complexity.  

From Table ~{\ref{tab:en-route_c2}}, we observe that LSTM-based models outperforms the Markov process-based model in Case 2. This result is reasonable since Markov process makes an assumption that future states depend only on the current state, not on the events that occurred before it. In other words, the state transition probability is independent of time which leads to prediction bias since Markov model can not capture the temporal patterns of uncertainties from weather.

\begin{table}[]
\caption{Quantitative Results (C1) of All the Methods for En-route Phase}
\begin{tabular}{lccccc}
\toprule
\multirow{2}{*}{Model} & \multirow{2}{*}{ADE} & \multirow{2}{*}{FDE} & \multicolumn{3}{c}{MAE}                              \\ \cmidrule{4-6} 
                       &                      &                      & longitude       & latitude        & altitude         \\ \midrule
V-LSTM                 & 842.7844             & 1477.2425             & 0.0061          &  0.0031         & 131.5484          \\
KF                     & 874.8100             & 1589.2196            & 0.0065          & 0.0035          & 95.4739          \\
HMM                    & \textbf{825.1741}             & \textbf{1354.3221}            & \textbf{0.0058}          & \textbf{0.0029}          & 113.8461         \\
Our                    & 1157.4851    & 2948.7744    & 0.0079 & 0.0059 & \textbf{91.6638} \\ \bottomrule
\end{tabular}
\label{tab:en-route_c1}
\end{table}

\begin{table}[]
\caption{Quantitative Results (C2) of All the Methods for En-route Phase}
\begin{tabular}{lccccc}
\toprule
\multirow{2}{*}{Model} & \multirow{2}{*}{ADE} & \multirow{2}{*}{FDE} & \multicolumn{3}{c}{MAE}                              \\ \cmidrule{4-6} 
                       &                      &                      & longitude       & latitude        & altitude         \\ \midrule
V-LSTM                 & 400.5694             & 769.4510             & 0.0030          & 0.0024          & 57.9083          \\
HMM                    & 572.8794             & 985.4811            & 0.0049          & 0.0028          & 97.5547         \\
Our                    & \textbf{228.6174}    & \textbf{359.3068}    & \textbf{0.0018} & \textbf{0.0013} & \textbf{23.4769} \\ \bottomrule
\end{tabular}
\label{tab:en-route_c2}
\end{table}

\subsection{Ablation Study} 
To comprehensive evaluate the effectiveness of our proposed modules, we further perform ablation studies. We remove a single component in an ablation study to construct new structure. We investigate four variants of our models as follows.
\begin{itemize}
    \item \textbf{Structural LSTM without Spatial Attention (SWSA):} This variant dose not have the spatial attention module. We concatenate hidden states of edgeLSTMs and the feature vector of the node directly as the input to nodeLSTMs.
    \item \textbf{Structural LSTM without Motion Constraints (SWMC):} This variant does have the motion constraints in the inference process. The predicted result depends on on-time sampling.
    \item \textbf{Encoder-Decoder without Dual-stage Attention (EDWDA):} This variant can be thought as a naive sequence-to-sequence structure. The intermediate (encoder) vector is directly fed into decoder.
    \item \textbf{Encoder-Decoder with Position-only Loss Function (EDWP):} The loss function used in this variant is to minimize the distance between predicted way-points and groundtruth without taking into account the aircraft's aerodynamics such as speed and acceleration.
\end{itemize}



In Table \ref{tab:takeoff_as} and Table \ref{tab:approach_as}, we observe our proposed motion constraint mechanism plays a more important role than other components because of the dramatically increased prediction errors. Besides, the introduction of attention module also greatly improves prediction accuracy but it seems that weather information does not make sense in the takeoff phase. We think the reason is that departure flights will be canceled or delayed when the weather is pretty bad but these ground trajectories are not considered in our prediction framework. In other words, the trajectories in takeoff phase, which are used to be trained or predicted, are collected when aircraft operate under normal weather conditions.

\begin{table}[htbp]
\caption{Quantitative Results of Ablation Study for Takeoff Phase}
\begin{tabular}{lccccc}
\toprule
\multirow{2}{*}{Model} & \multirow{2}{*}{ADE} & \multirow{2}{*}{FDE} & \multicolumn{3}{c}{MAE}                              \\ \cmidrule{4-6} 
                       &                      &                      & longitude       & latitude        & altitude         \\ \midrule
SWSA            & 840.8854             &1956.4782     & 0.0069        & 0.0044         & 180.3561       \\
SWMC        & 1453.5596     & 2561.5495          & 0.0124          & 0.0073          & 203.7228         \\
Our     & \textbf{312.3486}    & \textbf{437.8141}    & \textbf{0.0023} & \textbf{0.0019} & \textbf{54.8213} \\ \bottomrule
\end{tabular}
\label{tab:takeoff_as}
\end{table}

\begin{table}[htbp]
\caption{Quantitative Results of Ablation Study for Approach Phase}
\begin{tabular}{lccccc}
\toprule
\multirow{2}{*}{Model} & \multirow{2}{*}{ADE} & \multirow{2}{*}{FDE} & \multicolumn{3}{c}{MAE}                              \\ \cmidrule{4-6} 
                       &                      &                      & longitude       & latitude        & altitude         \\ \midrule
SWSA &  1165.7009        &  2241.7800          &0.0094           &0.0065           & 197.4589        \\
SWMC & 2264.8614           & 4158.4823         & 0.0187         & 0.0123         & 224.2156        \\
Our   & \textbf{450.9431}    & \textbf{766.2355}    & \textbf{0.0036} & \textbf{0.0024} & \textbf{92.8776} \\ \bottomrule
\end{tabular}
\label{tab:approach_as}
\end{table}

Table \ref{tab:en-route_as} shows the results of ablation study for en-route phase. Different from takeoff and approach, weather information is essential in the en-route trajectory prediction. The lack of aerodynamic information in loss function mainly affects the prediction accuracy of altitude and then affects the ADE and FED. Moreover, although we observe that the dual-stage attention module does not contribute alot to improve the prediction accuracy of coordinates in terms of MSE of latitude, longitude, and altitude, it helps reduce the prediction error of the entire predicted trajectory.

\begin{table}[htbp]
\caption{Quantitative Results of Ablation Study for En-route Phase}
\begin{tabular}{lccccc}
\toprule
\multirow{2}{*}{Model} & \multirow{2}{*}{ADE} & \multirow{2}{*}{FDE} & \multicolumn{3}{c}{MAE}                              \\ \cmidrule{4-6} 
                &                      &                      & longitude       & latitude  & altitude         \\ \midrule
EDWDA   & 320.0796             & 689.1445            & 0.0024          & 0.0018          & 68.5896          \\
EDWP        & 339.5309           & 733.5141           & 0.0024         & 0.0020        & 124.1873         \\
Our                    & \textbf{228.6174}    & \textbf{359.3068}    & \textbf{0.0018} & \textbf{0.0013} & \textbf{23.4769} \\ \bottomrule
\end{tabular}
\label{tab:en-route_as}
\end{table}


\section{Conclusion} \label{sec:conclusion}
In this article, we present the first flight trajectory prediction framework considering aircraft-aircraft interactions and flight phase characteristics by leveraging real-world big aviation data. Compared with existing flight trajectory prediction research based on historical trajectories, this paper takes uncertainties and aerodynamics into account. We first propose an structural LSTM with physical constraints, to learn latent spatio-temporal dynamics of aircraft itself and their interactions. Then we propose a method of representation of high-dimensional weather. The compact representation of all information is tweaked to learned by the dual-attention mechanism by minimizing aerodynamics-related loss function LVA. From the experimental results on our dataset, we show the superiority of our proposed approach and the effectiveness of each of the components. 
Our approach can facilitate effective air traffic management, which is of great importance to advance aerial mobility.

\begin{acks}
This research was done when authors (Zhang and Chen) were Research Assistant at Embry-Riddle Aeronautical University.
We thank faculty working in the Next-Generation Advanced Research (NEAR) Lab at Embry-Riddle Aeronautical University for providing collected aerial data.
\end{acks}

\bibliographystyle{ACM-Reference-Format}
\bibliography{sample-base}


\begin{thebibliography}{57}


\ifx \showCODEN    \undefined \def \showCODEN     #1{\unskip}     \fi
\ifx \showDOI      \undefined \def \showDOI       #1{#1}\fi
\ifx \showISBNx    \undefined \def \showISBNx     #1{\unskip}     \fi
\ifx \showISBNxiii \undefined \def \showISBNxiii  #1{\unskip}     \fi
\ifx \showISSN     \undefined \def \showISSN      #1{\unskip}     \fi
\ifx \showLCCN     \undefined \def \showLCCN      #1{\unskip}     \fi
\ifx \shownote     \undefined \def \shownote      #1{#1}          \fi
\ifx \showarticletitle \undefined \def \showarticletitle #1{#1}   \fi
\ifx \showURL      \undefined \def \showURL       {\relax}        \fi
\providecommand\bibfield[2]{#2}
\providecommand\bibinfo[2]{#2}
\providecommand\natexlab[1]{#1}
\providecommand\showeprint[2][]{arXiv:#2}

\bibitem[\protect\citeauthoryear{Administration}{Administration}{2009}]%
        {federal2009pilot}
\bibfield{author}{\bibinfo{person}{Federal~Aviation Administration}.}
  \bibinfo{year}{2009}\natexlab{}.
\newblock \bibinfo{booktitle}{\emph{Pilot's handbook of aeronautical
  knowledge}}.
\newblock \bibinfo{publisher}{Skyhorse Publishing Inc.}
\newblock


\bibitem[\protect\citeauthoryear{Alahi, Goel, Ramanathan, Robicquet, Fei-Fei,
  and Savarese}{Alahi et~al\mbox{.}}{2016}]%
        {alahi2016social}
\bibfield{author}{\bibinfo{person}{Alexandre Alahi}, \bibinfo{person}{Kratarth
  Goel}, \bibinfo{person}{Vignesh Ramanathan}, \bibinfo{person}{Alexandre
  Robicquet}, \bibinfo{person}{Li Fei-Fei}, {and} \bibinfo{person}{Silvio
  Savarese}.} \bibinfo{year}{2016}\natexlab{}.
\newblock \showarticletitle{Social lstm: Human trajectory prediction in crowded
  spaces}. In \bibinfo{booktitle}{\emph{Proceedings of the IEEE conference on
  computer vision and pattern recognition}}. \bibinfo{pages}{961--971}.
\newblock


\bibitem[\protect\citeauthoryear{An, Chen, Park, and Subrahmanian}{An
  et~al\mbox{.}}{2016}]%
        {an2016map}
\bibfield{author}{\bibinfo{person}{Bo An}, \bibinfo{person}{Haipeng Chen},
  \bibinfo{person}{Noseong Park}, {and} \bibinfo{person}{VS Subrahmanian}.}
  \bibinfo{year}{2016}\natexlab{}.
\newblock \showarticletitle{MAP: Frequency-based maximization of airline
  profits based on an ensemble forecasting approach}. In
  \bibinfo{booktitle}{\emph{Proceedings of the 22nd ACM SIGKDD International
  Conference on Knowledge Discovery and Data Mining}}.
  \bibinfo{pages}{421--430}.
\newblock


\bibitem[\protect\citeauthoryear{Ayhan and Samet}{Ayhan and Samet}{2016}]%
        {ayhan2016aircraft}
\bibfield{author}{\bibinfo{person}{Samet Ayhan} {and} \bibinfo{person}{Hanan
  Samet}.} \bibinfo{year}{2016}\natexlab{}.
\newblock \showarticletitle{Aircraft trajectory prediction made easy with
  predictive analytics}. In \bibinfo{booktitle}{\emph{Proceedings of the 22nd
  ACM SIGKDD International Conference on Knowledge Discovery and Data Mining}}.
  \bibinfo{pages}{21--30}.
\newblock


\bibitem[\protect\citeauthoryear{Bai, Yao, Kanhere, Wang, Liu, and Yang}{Bai
  et~al\mbox{.}}{2019}]%
        {bai2019spatio}
\bibfield{author}{\bibinfo{person}{Lei Bai}, \bibinfo{person}{Lina Yao},
  \bibinfo{person}{Salil~S Kanhere}, \bibinfo{person}{Xianzhi Wang},
  \bibinfo{person}{Wei Liu}, {and} \bibinfo{person}{Zheng Yang}.}
  \bibinfo{year}{2019}\natexlab{}.
\newblock \showarticletitle{Spatio-temporal graph convolutional and recurrent
  networks for citywide passenger demand prediction}. In
  \bibinfo{booktitle}{\emph{Proceedings of the 28th ACM International
  Conference on Information and Knowledge Management}}.
  \bibinfo{pages}{2293--2296}.
\newblock


\bibitem[\protect\citeauthoryear{Barratt, Kochenderfer, and Boyd}{Barratt
  et~al\mbox{.}}{2018}]%
        {barratt2018learning}
\bibfield{author}{\bibinfo{person}{Shane~T Barratt}, \bibinfo{person}{Mykel~J
  Kochenderfer}, {and} \bibinfo{person}{Stephen~P Boyd}.}
  \bibinfo{year}{2018}\natexlab{}.
\newblock \showarticletitle{Learning probabilistic trajectory models of
  aircraft in terminal airspace from position data}.
\newblock \bibinfo{journal}{\emph{IEEE Transactions on Intelligent
  Transportation Systems}} \bibinfo{volume}{20}, \bibinfo{number}{9}
  (\bibinfo{year}{2018}), \bibinfo{pages}{3536--3545}.
\newblock


\bibitem[\protect\citeauthoryear{Basora, Morio, and Mailhot}{Basora
  et~al\mbox{.}}{2017}]%
        {basora2017trajectory}
\bibfield{author}{\bibinfo{person}{Luis Basora},
  \bibinfo{person}{J{\'e}r{\^o}me Morio}, {and} \bibinfo{person}{Corentin
  Mailhot}.} \bibinfo{year}{2017}\natexlab{}.
\newblock \showarticletitle{A trajectory clustering framework to analyse air
  traffic flows}.
\newblock \bibinfo{journal}{\emph{7th SESAR Innovation Days}}
  (\bibinfo{year}{2017}), \bibinfo{pages}{1--8}.
\newblock


\bibitem[\protect\citeauthoryear{Campello, Moulavi, and Sander}{Campello
  et~al\mbox{.}}{2013}]%
        {campello2013density}
\bibfield{author}{\bibinfo{person}{Ricardo~JGB Campello},
  \bibinfo{person}{Davoud Moulavi}, {and} \bibinfo{person}{J{\"o}rg Sander}.}
  \bibinfo{year}{2013}\natexlab{}.
\newblock \showarticletitle{Density-based clustering based on hierarchical
  density estimates}. In \bibinfo{booktitle}{\emph{Pacific-Asia conference on
  knowledge discovery and data mining}}. Springer, \bibinfo{pages}{160--172}.
\newblock


\bibitem[\protect\citeauthoryear{Chandra, Bhattacharya, Bera, and
  Manocha}{Chandra et~al\mbox{.}}{2019}]%
        {chandra2019traphic}
\bibfield{author}{\bibinfo{person}{Rohan Chandra}, \bibinfo{person}{Uttaran
  Bhattacharya}, \bibinfo{person}{Aniket Bera}, {and} \bibinfo{person}{Dinesh
  Manocha}.} \bibinfo{year}{2019}\natexlab{}.
\newblock \showarticletitle{Traphic: Trajectory prediction in dense and
  heterogeneous traffic using weighted interactions}. In
  \bibinfo{booktitle}{\emph{Proceedings of the IEEE/CVF Conference on Computer
  Vision and Pattern Recognition}}. \bibinfo{pages}{8483--8492}.
\newblock


\bibitem[\protect\citeauthoryear{Chung, Gulcehre, Cho, and Bengio}{Chung
  et~al\mbox{.}}{2014}]%
        {chung2014empirical}
\bibfield{author}{\bibinfo{person}{Junyoung Chung}, \bibinfo{person}{Caglar
  Gulcehre}, \bibinfo{person}{KyungHyun Cho}, {and} \bibinfo{person}{Yoshua
  Bengio}.} \bibinfo{year}{2014}\natexlab{}.
\newblock \showarticletitle{Empirical evaluation of gated recurrent neural
  networks on sequence modeling}.
\newblock \bibinfo{journal}{\emph{arXiv preprint arXiv:1412.3555}}
  (\bibinfo{year}{2014}).
\newblock


\bibitem[\protect\citeauthoryear{Churchill and Bloem}{Churchill and
  Bloem}{2019}]%
        {churchill2019clustering}
\bibfield{author}{\bibinfo{person}{Andrew~M Churchill} {and}
  \bibinfo{person}{Michael Bloem}.} \bibinfo{year}{2019}\natexlab{}.
\newblock \showarticletitle{Clustering aircraft trajectories on the airport
  surface}. In \bibinfo{booktitle}{\emph{Proceedings of the 13th USA/Europe Air
  Traffic Management Research and Development Seminar, Chicago, IL, USA}}.
  \bibinfo{pages}{10--13}.
\newblock


\bibitem[\protect\citeauthoryear{FAA}{FAA}{2021}]%
        {waypoints}
\bibfield{author}{\bibinfo{person}{FAA}.} \bibinfo{year}{2021}\natexlab{}.
\newblock \bibinfo{booktitle}{\emph{Fixes/Waypoints}}.
\newblock
\urldef\tempurl%
\url{https://www.faa.gov/air_traffic/flight_info/aeronav/aero_data/loc_id_search/Fixes_Waypoints/}
\showURL{%
Retrieved October 4, 2020 from \tempurl}


\bibitem[\protect\citeauthoryear{Fang, Lin, Yang, Yu, and Xu}{Fang
  et~al\mbox{.}}{2019}]%
        {fang2019citytracker}
\bibfield{author}{\bibinfo{person}{Shih-Hau Fang}, \bibinfo{person}{Larry Lin},
  \bibinfo{person}{Yi-Ting Yang}, \bibinfo{person}{Xiaotong Yu}, {and}
  \bibinfo{person}{Zhezhuang Xu}.} \bibinfo{year}{2019}\natexlab{}.
\newblock \showarticletitle{CityTracker: Citywide Individual and Crowd
  Trajectory Analysis Using Hidden Markov Model}.
\newblock \bibinfo{journal}{\emph{IEEE Sensors Journal}} \bibinfo{volume}{19},
  \bibinfo{number}{17} (\bibinfo{year}{2019}), \bibinfo{pages}{7693--7701}.
\newblock


\bibitem[\protect\citeauthoryear{Gallego, Comendador, Nieto, Imaz, and
  Vald{\'e}s}{Gallego et~al\mbox{.}}{2018}]%
        {gallego2018analysis}
\bibfield{author}{\bibinfo{person}{Christian Eduardo~Verdonk Gallego},
  \bibinfo{person}{V{\'\i}ctor Fernando~G{\'o}mez Comendador},
  \bibinfo{person}{Francisco Javier~S{\'a}ez Nieto},
  \bibinfo{person}{Guillermo~Orenga Imaz}, {and} \bibinfo{person}{Rosa
  Mar{\'\i}a~Arnaldo Vald{\'e}s}.} \bibinfo{year}{2018}\natexlab{}.
\newblock \showarticletitle{Analysis of air traffic control operational impact
  on aircraft vertical profiles supported by machine learning}.
\newblock \bibinfo{journal}{\emph{Transportation research part C: emerging
  technologies}}  \bibinfo{volume}{95} (\bibinfo{year}{2018}),
  \bibinfo{pages}{883--903}.
\newblock


\bibitem[\protect\citeauthoryear{Gariel, Srivastava, and Feron}{Gariel
  et~al\mbox{.}}{2011}]%
        {gariel2011trajectory}
\bibfield{author}{\bibinfo{person}{Maxime Gariel}, \bibinfo{person}{Ashok~N
  Srivastava}, {and} \bibinfo{person}{Eric Feron}.}
  \bibinfo{year}{2011}\natexlab{}.
\newblock \showarticletitle{Trajectory clustering and an application to
  airspace monitoring}.
\newblock \bibinfo{journal}{\emph{IEEE Transactions on Intelligent
  Transportation Systems}} \bibinfo{volume}{12}, \bibinfo{number}{4}
  (\bibinfo{year}{2011}), \bibinfo{pages}{1511--1524}.
\newblock


\bibitem[\protect\citeauthoryear{generation ERAU Applied
  Research~Lab}{generation ERAU Applied Research~Lab}{2020}]%
        {ERAUlive}
\bibfield{author}{\bibinfo{person}{Next generation ERAU Applied Research~Lab}.}
  \bibinfo{year}{2020}\natexlab{}.
\newblock \bibinfo{booktitle}{\emph{ERAU Traffic Live}}.
\newblock
\urldef\tempurl%
\url{https://www.near.aero/erau-livetraffic}
\showURL{%
Retrieved October 1, 2020 from \tempurl}


\bibitem[\protect\citeauthoryear{Graves}{Graves}{2013}]%
        {graves2013generating}
\bibfield{author}{\bibinfo{person}{Alex Graves}.}
  \bibinfo{year}{2013}\natexlab{}.
\newblock \showarticletitle{Generating sequences with recurrent neural
  networks}.
\newblock \bibinfo{journal}{\emph{arXiv preprint arXiv:1308.0850}}
  (\bibinfo{year}{2013}).
\newblock


\bibitem[\protect\citeauthoryear{Gui, Zhang, and Peng}{Gui
  et~al\mbox{.}}{2021}]%
        {xuhao2021trajectory}
\bibfield{author}{\bibinfo{person}{Xuhao Gui}, \bibinfo{person}{Junfeng Zhang},
  {and} \bibinfo{person}{Zihan Peng}.} \bibinfo{year}{2021}\natexlab{}.
\newblock \showarticletitle{Trajectory clustering for arrival aircraft via new
  trajectory representation}.
\newblock \bibinfo{journal}{\emph{Journal of Systems Engineering and
  Electronics}} \bibinfo{volume}{32}, \bibinfo{number}{2}
  (\bibinfo{year}{2021}), \bibinfo{pages}{473--486}.
\newblock


\bibitem[\protect\citeauthoryear{Gupta, Johnson, Fei-Fei, Savarese, and
  Alahi}{Gupta et~al\mbox{.}}{2018}]%
        {gupta2018social}
\bibfield{author}{\bibinfo{person}{Agrim Gupta}, \bibinfo{person}{Justin
  Johnson}, \bibinfo{person}{Li Fei-Fei}, \bibinfo{person}{Silvio Savarese},
  {and} \bibinfo{person}{Alexandre Alahi}.} \bibinfo{year}{2018}\natexlab{}.
\newblock \showarticletitle{Social gan: Socially acceptable trajectories with
  generative adversarial networks}. In \bibinfo{booktitle}{\emph{Proceedings of
  the IEEE Conference on Computer Vision and Pattern Recognition}}.
  \bibinfo{pages}{2255--2264}.
\newblock


\bibitem[\protect\citeauthoryear{Habler and Shabtai}{Habler and
  Shabtai}{2018}]%
        {habler2018using}
\bibfield{author}{\bibinfo{person}{Edan Habler} {and} \bibinfo{person}{Asaf
  Shabtai}.} \bibinfo{year}{2018}\natexlab{}.
\newblock \showarticletitle{Using LSTM encoder-decoder algorithm for detecting
  anomalous ADS-B messages}.
\newblock \bibinfo{journal}{\emph{Computers \& Security}}  \bibinfo{volume}{78}
  (\bibinfo{year}{2018}), \bibinfo{pages}{155--173}.
\newblock


\bibitem[\protect\citeauthoryear{Helbing and Molnar}{Helbing and
  Molnar}{1995}]%
        {helbing1995social}
\bibfield{author}{\bibinfo{person}{Dirk Helbing} {and} \bibinfo{person}{Peter
  Molnar}.} \bibinfo{year}{1995}\natexlab{}.
\newblock \showarticletitle{Social force model for pedestrian dynamics}.
\newblock \bibinfo{journal}{\emph{Physical review E}} \bibinfo{volume}{51},
  \bibinfo{number}{5} (\bibinfo{year}{1995}), \bibinfo{pages}{4282}.
\newblock


\bibitem[\protect\citeauthoryear{Hochreiter and Schmidhuber}{Hochreiter and
  Schmidhuber}{1997}]%
        {hochreiter1997long}
\bibfield{author}{\bibinfo{person}{Sepp Hochreiter} {and}
  \bibinfo{person}{J{\"u}rgen Schmidhuber}.} \bibinfo{year}{1997}\natexlab{}.
\newblock \showarticletitle{Long short-term memory}.
\newblock \bibinfo{journal}{\emph{Neural computation}} \bibinfo{volume}{9},
  \bibinfo{number}{8} (\bibinfo{year}{1997}), \bibinfo{pages}{1735--1780}.
\newblock


\bibitem[\protect\citeauthoryear{Jain, Zamir, Savarese, and Saxena}{Jain
  et~al\mbox{.}}{2016}]%
        {jain2016structural}
\bibfield{author}{\bibinfo{person}{Ashesh Jain}, \bibinfo{person}{Amir~R
  Zamir}, \bibinfo{person}{Silvio Savarese}, {and} \bibinfo{person}{Ashutosh
  Saxena}.} \bibinfo{year}{2016}\natexlab{}.
\newblock \showarticletitle{Structural-rnn: Deep learning on spatio-temporal
  graphs}. In \bibinfo{booktitle}{\emph{Proceedings of the ieee conference on
  computer vision and pattern recognition}}. \bibinfo{pages}{5308--5317}.
\newblock


\bibitem[\protect\citeauthoryear{Ju, Wang, Long, Zhang, and Chang}{Ju
  et~al\mbox{.}}{2019}]%
        {ju2019interaction}
\bibfield{author}{\bibinfo{person}{Ce Ju}, \bibinfo{person}{Zheng Wang},
  \bibinfo{person}{Cheng Long}, \bibinfo{person}{Xiaoyu Zhang}, {and}
  \bibinfo{person}{Dong~Eui Chang}.} \bibinfo{year}{2019}\natexlab{}.
\newblock \showarticletitle{Interaction-aware kalman neural networks for
  trajectory prediction}. In \bibinfo{booktitle}{\emph{2020 IEEE Intelligent
  Vehicles Symposium (IV)}}. IEEE, \bibinfo{pages}{1793--1800}.
\newblock


\bibitem[\protect\citeauthoryear{Jung, Hong, and Lee}{Jung
  et~al\mbox{.}}{2018}]%
        {jung2018data}
\bibfield{author}{\bibinfo{person}{Soyeon Jung}, \bibinfo{person}{Sungkweon
  Hong}, {and} \bibinfo{person}{Keumjin Lee}.} \bibinfo{year}{2018}\natexlab{}.
\newblock \showarticletitle{A Data-Driven Air Traffic Sequencing Model Based on
  Pairwise Preference Learning}.
\newblock \bibinfo{journal}{\emph{IEEE Transactions on Intelligent
  Transportation Systems}} \bibinfo{volume}{20}, \bibinfo{number}{3}
  (\bibinfo{year}{2018}), \bibinfo{pages}{803--816}.
\newblock


\bibitem[\protect\citeauthoryear{Khodayar and Wang}{Khodayar and Wang}{2018}]%
        {khodayar2018spatio}
\bibfield{author}{\bibinfo{person}{Mahdi Khodayar} {and}
  \bibinfo{person}{Jianhui Wang}.} \bibinfo{year}{2018}\natexlab{}.
\newblock \showarticletitle{Spatio-temporal graph deep neural network for
  short-term wind speed forecasting}.
\newblock \bibinfo{journal}{\emph{IEEE Transactions on Sustainable Energy}}
  \bibinfo{volume}{10}, \bibinfo{number}{2} (\bibinfo{year}{2018}),
  \bibinfo{pages}{670--681}.
\newblock


\bibitem[\protect\citeauthoryear{Laboratories}{Laboratories}{2021}]%
        {RAP}
\bibfield{author}{\bibinfo{person}{NOAA Earth System~Research Laboratories}.}
  \bibinfo{year}{2021}\natexlab{}.
\newblock \bibinfo{booktitle}{\emph{Rapid Refresh (RAP)}}.
\newblock
\urldef\tempurl%
\url{https://rapidrefresh.noaa.gov/}
\showURL{%
Retrieved October 1, 2020 from \tempurl}


\bibitem[\protect\citeauthoryear{Lee, Han, and Whang}{Lee
  et~al\mbox{.}}{2007}]%
        {lee2007trajectory}
\bibfield{author}{\bibinfo{person}{Jae-Gil Lee}, \bibinfo{person}{Jiawei Han},
  {and} \bibinfo{person}{Kyu-Young Whang}.} \bibinfo{year}{2007}\natexlab{}.
\newblock \showarticletitle{Trajectory clustering: a partition-and-group
  framework}. In \bibinfo{booktitle}{\emph{Proceedings of the 2007 ACM SIGMOD
  international conference on Management of data}}. \bibinfo{pages}{593--604}.
\newblock


\bibitem[\protect\citeauthoryear{Lee, Choi, Vernaza, Choy, Torr, and
  Chandraker}{Lee et~al\mbox{.}}{2017}]%
        {lee2017desire}
\bibfield{author}{\bibinfo{person}{Namhoon Lee}, \bibinfo{person}{Wongun Choi},
  \bibinfo{person}{Paul Vernaza}, \bibinfo{person}{Christopher~B Choy},
  \bibinfo{person}{Philip~HS Torr}, {and} \bibinfo{person}{Manmohan
  Chandraker}.} \bibinfo{year}{2017}\natexlab{}.
\newblock \showarticletitle{Desire: Distant future prediction in dynamic scenes
  with interacting agents}. In \bibinfo{booktitle}{\emph{Proceedings of the
  IEEE Conference on Computer Vision and Pattern Recognition}}.
  \bibinfo{pages}{336--345}.
\newblock


\bibitem[\protect\citeauthoryear{Leonardi}{Leonardi}{2018}]%
        {leonardi2018ads}
\bibfield{author}{\bibinfo{person}{Mauro Leonardi}.}
  \bibinfo{year}{2018}\natexlab{}.
\newblock \showarticletitle{ADS-B anomalies and intrusions detection by sensor
  clocks tracking}.
\newblock \bibinfo{journal}{\emph{IEEE Trans. Aerospace Electron. Systems}}
  \bibinfo{volume}{55}, \bibinfo{number}{5} (\bibinfo{year}{2018}),
  \bibinfo{pages}{2370--2381}.
\newblock


\bibitem[\protect\citeauthoryear{Lin, Zhang, and Liu}{Lin
  et~al\mbox{.}}{2018}]%
        {lin2018algorithm}
\bibfield{author}{\bibinfo{person}{Yi Lin}, \bibinfo{person}{Jian-wei Zhang},
  {and} \bibinfo{person}{Hong Liu}.} \bibinfo{year}{2018}\natexlab{}.
\newblock \showarticletitle{An algorithm for trajectory prediction of flight
  plan based on relative motion between positions}.
\newblock \bibinfo{journal}{\emph{Frontiers of Information Technology \&
  Electronic Engineering}} \bibinfo{volume}{19}, \bibinfo{number}{7}
  (\bibinfo{year}{2018}), \bibinfo{pages}{905--916}.
\newblock


\bibitem[\protect\citeauthoryear{Lin, Zhang, and Liu}{Lin
  et~al\mbox{.}}{2019}]%
        {lin2019deep}
\bibfield{author}{\bibinfo{person}{Yi Lin}, \bibinfo{person}{Jian-wei Zhang},
  {and} \bibinfo{person}{Hong Liu}.} \bibinfo{year}{2019}\natexlab{}.
\newblock \showarticletitle{Deep learning based short-term air traffic flow
  prediction considering temporal--spatial correlation}.
\newblock \bibinfo{journal}{\emph{Aerospace Science and Technology}}
  \bibinfo{volume}{93} (\bibinfo{year}{2019}), \bibinfo{pages}{105113}.
\newblock


\bibitem[\protect\citeauthoryear{Liu and Hwang}{Liu and Hwang}{2011}]%
        {liu2011probabilistic}
\bibfield{author}{\bibinfo{person}{Weiyi Liu} {and} \bibinfo{person}{Inseok
  Hwang}.} \bibinfo{year}{2011}\natexlab{}.
\newblock \showarticletitle{Probabilistic trajectory prediction and conflict
  detection for air traffic control}.
\newblock \bibinfo{journal}{\emph{Journal of Guidance, Control, and Dynamics}}
  \bibinfo{volume}{34}, \bibinfo{number}{6} (\bibinfo{year}{2011}),
  \bibinfo{pages}{1779--1789}.
\newblock


\bibitem[\protect\citeauthoryear{Liu and Hansen}{Liu and Hansen}{2018}]%
        {liu2018predicting}
\bibfield{author}{\bibinfo{person}{Yulin Liu} {and} \bibinfo{person}{Mark
  Hansen}.} \bibinfo{year}{2018}\natexlab{}.
\newblock \showarticletitle{Predicting aircraft trajectories: a deep generative
  convolutional recurrent neural networks approach}.
\newblock \bibinfo{journal}{\emph{arXiv preprint arXiv:1812.11670}}
  (\bibinfo{year}{2018}).
\newblock


\bibitem[\protect\citeauthoryear{Nikhil and Tran~Morris}{Nikhil and
  Tran~Morris}{2018}]%
        {nikhil2018convolutional}
\bibfield{author}{\bibinfo{person}{Nishant Nikhil} {and}
  \bibinfo{person}{Brendan Tran~Morris}.} \bibinfo{year}{2018}\natexlab{}.
\newblock \showarticletitle{Convolutional neural network for trajectory
  prediction}. In \bibinfo{booktitle}{\emph{Proceedings of the European
  Conference on Computer Vision (ECCV) Workshops}}. \bibinfo{pages}{0--0}.
\newblock


\bibitem[\protect\citeauthoryear{of~Transportation~Statistics}{of~Transportation~Statistics}{2021}]%
        {AOTP}
\bibfield{author}{\bibinfo{person}{Bureau of Transportation~Statistics}.}
  \bibinfo{year}{2021}\natexlab{}.
\newblock \bibinfo{booktitle}{\emph{Airline On-Time Performance Database}}.
\newblock
\urldef\tempurl%
\url{https://www.transtats.bts.gov/Tables.asp?DB_ID=120}
\showURL{%
Retrieved October 1, 2020 from \tempurl}


\bibitem[\protect\citeauthoryear{Olive and Basora}{Olive and Basora}{2020}]%
        {olive2020detection}
\bibfield{author}{\bibinfo{person}{Xavier Olive} {and} \bibinfo{person}{Luis
  Basora}.} \bibinfo{year}{2020}\natexlab{}.
\newblock \showarticletitle{Detection and identification of significant events
  in historical aircraft trajectory data}.
\newblock \bibinfo{journal}{\emph{Transportation Research Part C: Emerging
  Technologies}}  \bibinfo{volume}{119} (\bibinfo{year}{2020}),
  \bibinfo{pages}{102737}.
\newblock


\bibitem[\protect\citeauthoryear{Olive and Morio}{Olive and Morio}{2019}]%
        {olive2019trajectory}
\bibfield{author}{\bibinfo{person}{Xavier Olive} {and}
  \bibinfo{person}{J{\'e}r{\^o}me Morio}.} \bibinfo{year}{2019}\natexlab{}.
\newblock \showarticletitle{Trajectory clustering of air traffic flows around
  airports}.
\newblock \bibinfo{journal}{\emph{Aerospace Science and Technology}}
  \bibinfo{volume}{84} (\bibinfo{year}{2019}), \bibinfo{pages}{776--781}.
\newblock


\bibitem[\protect\citeauthoryear{Pang and Liu}{Pang and Liu}{2020}]%
        {pang2020conditional}
\bibfield{author}{\bibinfo{person}{Yutian Pang} {and} \bibinfo{person}{Yongming
  Liu}.} \bibinfo{year}{2020}\natexlab{}.
\newblock \showarticletitle{Conditional Generative Adversarial Networks (CGAN)
  for Aircraft Trajectory Prediction considering weather effects}. In
  \bibinfo{booktitle}{\emph{AIAA Scitech 2020 Forum}}. \bibinfo{pages}{1853}.
\newblock


\bibitem[\protect\citeauthoryear{Pang, Yao, Hu, and Liu}{Pang
  et~al\mbox{.}}{2019}]%
        {pang2019recurrent}
\bibfield{author}{\bibinfo{person}{Yutian Pang}, \bibinfo{person}{Houpu Yao},
  \bibinfo{person}{Jueming Hu}, {and} \bibinfo{person}{Yongming Liu}.}
  \bibinfo{year}{2019}\natexlab{}.
\newblock \showarticletitle{A Recurrent Neural Network Approach for Aircraft
  Trajectory Prediction with Weather Features From Sherlock}. In
  \bibinfo{booktitle}{\emph{AIAA Aviation 2019 Forum}}. \bibinfo{pages}{3413}.
\newblock


\bibitem[\protect\citeauthoryear{Pathirana, Savkin, and Jha}{Pathirana
  et~al\mbox{.}}{2003}]%
        {pathirana2003mobility}
\bibfield{author}{\bibinfo{person}{Pubudu~N Pathirana},
  \bibinfo{person}{Andrey~V Savkin}, {and} \bibinfo{person}{Sanjay Jha}.}
  \bibinfo{year}{2003}\natexlab{}.
\newblock \showarticletitle{Mobility modelling and trajectory prediction for
  cellular networks with mobile base stations}. In
  \bibinfo{booktitle}{\emph{Proceedings of the 4th ACM international symposium
  on Mobile ad hoc networking \& computing}}. \bibinfo{pages}{213--221}.
\newblock


\bibitem[\protect\citeauthoryear{Qiao, Shen, Wang, Han, and Zhu}{Qiao
  et~al\mbox{.}}{2014}]%
        {qiao2014self}
\bibfield{author}{\bibinfo{person}{Shaojie Qiao}, \bibinfo{person}{Dayong
  Shen}, \bibinfo{person}{Xiaoteng Wang}, \bibinfo{person}{Nan Han}, {and}
  \bibinfo{person}{William Zhu}.} \bibinfo{year}{2014}\natexlab{}.
\newblock \showarticletitle{A self-adaptive parameter selection trajectory
  prediction approach via hidden Markov models}.
\newblock \bibinfo{journal}{\emph{IEEE Transactions on Intelligent
  Transportation Systems}} \bibinfo{volume}{16}, \bibinfo{number}{1}
  (\bibinfo{year}{2014}), \bibinfo{pages}{284--296}.
\newblock


\bibitem[\protect\citeauthoryear{Qin, Song, Chen, Cheng, Jiang, and
  Cottrell}{Qin et~al\mbox{.}}{2017}]%
        {qin2017dual}
\bibfield{author}{\bibinfo{person}{Yao Qin}, \bibinfo{person}{Dongjin Song},
  \bibinfo{person}{Haifeng Chen}, \bibinfo{person}{Wei Cheng},
  \bibinfo{person}{Guofei Jiang}, {and} \bibinfo{person}{Garrison Cottrell}.}
  \bibinfo{year}{2017}\natexlab{}.
\newblock \showarticletitle{A dual-stage attention-based recurrent neural
  network for time series prediction}.
\newblock \bibinfo{journal}{\emph{arXiv preprint arXiv:1704.02971}}
  (\bibinfo{year}{2017}).
\newblock


\bibitem[\protect\citeauthoryear{Sadeghian, Kosaraju, Sadeghian, Hirose,
  Rezatofighi, and Savarese}{Sadeghian et~al\mbox{.}}{2019}]%
        {sadeghian2019sophie}
\bibfield{author}{\bibinfo{person}{Amir Sadeghian}, \bibinfo{person}{Vineet
  Kosaraju}, \bibinfo{person}{Ali Sadeghian}, \bibinfo{person}{Noriaki Hirose},
  \bibinfo{person}{Hamid Rezatofighi}, {and} \bibinfo{person}{Silvio
  Savarese}.} \bibinfo{year}{2019}\natexlab{}.
\newblock \showarticletitle{Sophie: An attentive gan for predicting paths
  compliant to social and physical constraints}. In
  \bibinfo{booktitle}{\emph{Proceedings of the IEEE/CVF Conference on Computer
  Vision and Pattern Recognition}}. \bibinfo{pages}{1349--1358}.
\newblock


\bibitem[\protect\citeauthoryear{Shi, Xu, and Pan}{Shi et~al\mbox{.}}{2020}]%
        {shi20204}
\bibfield{author}{\bibinfo{person}{Zhiyuan Shi}, \bibinfo{person}{Min Xu},
  {and} \bibinfo{person}{Quan Pan}.} \bibinfo{year}{2020}\natexlab{}.
\newblock \showarticletitle{4-D Flight Trajectory Prediction With Constrained
  LSTM Network}.
\newblock \bibinfo{journal}{\emph{IEEE Transactions on Intelligent
  Transportation Systems}} (\bibinfo{year}{2020}).
\newblock


\bibitem[\protect\citeauthoryear{Shi, Xu, Pan, Yan, and Zhang}{Shi
  et~al\mbox{.}}{2018}]%
        {shi2018lstm}
\bibfield{author}{\bibinfo{person}{Zhiyuan Shi}, \bibinfo{person}{Min Xu},
  \bibinfo{person}{Quan Pan}, \bibinfo{person}{Bing Yan}, {and}
  \bibinfo{person}{Haimin Zhang}.} \bibinfo{year}{2018}\natexlab{}.
\newblock \showarticletitle{LSTM-based flight trajectory prediction}. In
  \bibinfo{booktitle}{\emph{2018 International Joint Conference on Neural
  Networks (IJCNN)}}. IEEE, \bibinfo{pages}{1--8}.
\newblock


\bibitem[\protect\citeauthoryear{Sun, Ellerbroek, and Hoekstra}{Sun
  et~al\mbox{.}}{2016}]%
        {sun2016large}
\bibfield{author}{\bibinfo{person}{Junzi Sun}, \bibinfo{person}{Joost
  Ellerbroek}, {and} \bibinfo{person}{Jacco Hoekstra}.}
  \bibinfo{year}{2016}\natexlab{}.
\newblock \showarticletitle{Large-scale flight phase identification from ads-b
  data using machine learning methods}. In \bibinfo{booktitle}{\emph{7th
  International Conference on Research in Air Transportation}}.
\newblock


\bibitem[\protect\citeauthoryear{Sun, Ellerbroek, and Hoekstra}{Sun
  et~al\mbox{.}}{2017}]%
        {sun2017flight}
\bibfield{author}{\bibinfo{person}{Junzi Sun}, \bibinfo{person}{Joost
  Ellerbroek}, {and} \bibinfo{person}{Jacco Hoekstra}.}
  \bibinfo{year}{2017}\natexlab{}.
\newblock \showarticletitle{Flight extraction and phase identification for
  large automatic dependent surveillance--broadcast datasets}.
\newblock \bibinfo{journal}{\emph{Journal of Aerospace Information Systems}}
  \bibinfo{volume}{14}, \bibinfo{number}{10} (\bibinfo{year}{2017}),
  \bibinfo{pages}{566--572}.
\newblock


\bibitem[\protect\citeauthoryear{Tabassum, Allen, and Semke}{Tabassum
  et~al\mbox{.}}{2017}]%
        {tabassum2017ads}
\bibfield{author}{\bibinfo{person}{Asma Tabassum}, \bibinfo{person}{Nicholas
  Allen}, {and} \bibinfo{person}{William Semke}.}
  \bibinfo{year}{2017}\natexlab{}.
\newblock \showarticletitle{ADS-B message contents evaluation and breakdown of
  anomalies}. In \bibinfo{booktitle}{\emph{2017 IEEE/AIAA 36th Digital Avionics
  Systems Conference (DASC)}}. IEEE, \bibinfo{pages}{1--8}.
\newblock


\bibitem[\protect\citeauthoryear{Trautman and Krause}{Trautman and
  Krause}{2010}]%
        {trautman2010unfreezing}
\bibfield{author}{\bibinfo{person}{Peter Trautman} {and}
  \bibinfo{person}{Andreas Krause}.} \bibinfo{year}{2010}\natexlab{}.
\newblock \showarticletitle{Unfreezing the robot: Navigation in dense,
  interacting crowds}. In \bibinfo{booktitle}{\emph{2010 IEEE/RSJ International
  Conference on Intelligent Robots and Systems}}. IEEE,
  \bibinfo{pages}{797--803}.
\newblock


\bibitem[\protect\citeauthoryear{Vaswani, Shazeer, Parmar, Uszkoreit, Jones,
  Gomez, Kaiser, and Polosukhin}{Vaswani et~al\mbox{.}}{2017}]%
        {vaswani2017attention}
\bibfield{author}{\bibinfo{person}{Ashish Vaswani}, \bibinfo{person}{Noam
  Shazeer}, \bibinfo{person}{Niki Parmar}, \bibinfo{person}{Jakob Uszkoreit},
  \bibinfo{person}{Llion Jones}, \bibinfo{person}{Aidan~N Gomez},
  \bibinfo{person}{{\L}ukasz Kaiser}, {and} \bibinfo{person}{Illia
  Polosukhin}.} \bibinfo{year}{2017}\natexlab{}.
\newblock \showarticletitle{Attention is all you need}. In
  \bibinfo{booktitle}{\emph{Advances in neural information processing
  systems}}. \bibinfo{pages}{5998--6008}.
\newblock


\bibitem[\protect\citeauthoryear{Vemula, Muelling, and Oh}{Vemula
  et~al\mbox{.}}{2018}]%
        {vemula2018social}
\bibfield{author}{\bibinfo{person}{Anirudh Vemula}, \bibinfo{person}{Katharina
  Muelling}, {and} \bibinfo{person}{Jean Oh}.} \bibinfo{year}{2018}\natexlab{}.
\newblock \showarticletitle{Social attention: Modeling attention in human
  crowds}. In \bibinfo{booktitle}{\emph{2018 IEEE international Conference on
  Robotics and Automation (ICRA)}}. IEEE, \bibinfo{pages}{4601--4607}.
\newblock


\bibitem[\protect\citeauthoryear{Wang and Zhao}{Wang and Zhao}{2020}]%
        {WANG2020103}
\bibfield{author}{\bibinfo{person}{Guoqing Wang} {and} \bibinfo{person}{Wenhao
  Zhao}.} \bibinfo{year}{2020}\natexlab{}.
\newblock \showarticletitle{Chapter 3 - The requirement organization of the
  avionics system}.
\newblock In \bibinfo{booktitle}{\emph{The Principles of Integrated Technology
  in Avionics Systems}}, \bibfield{editor}{\bibinfo{person}{Guoqing Wang} {and}
  \bibinfo{person}{Wenhao Zhao}} (Eds.). \bibinfo{publisher}{Academic Press},
  \bibinfo{pages}{103 -- 167}.
\newblock
\showISBNx{978-0-12-816651-2}
\urldef\tempurl%
\url{https://doi.org/10.1016/B978-0-12-816651-2.00003-4}
\showDOI{\tempurl}


\bibitem[\protect\citeauthoryear{Wang, Yin, Chen, Wo, Xu, and Zheng}{Wang
  et~al\mbox{.}}{2019}]%
        {wang2019origin}
\bibfield{author}{\bibinfo{person}{Yuandong Wang}, \bibinfo{person}{Hongzhi
  Yin}, \bibinfo{person}{Hongxu Chen}, \bibinfo{person}{Tianyu Wo},
  \bibinfo{person}{Jie Xu}, {and} \bibinfo{person}{Kai Zheng}.}
  \bibinfo{year}{2019}\natexlab{}.
\newblock \showarticletitle{Origin-destination matrix prediction via graph
  convolution: a new perspective of passenger demand modeling}. In
  \bibinfo{booktitle}{\emph{Proceedings of the 25th ACM SIGKDD International
  Conference on Knowledge Discovery \& Data Mining}}.
  \bibinfo{pages}{1227--1235}.
\newblock


\bibitem[\protect\citeauthoryear{Warnock-Smith, O'Connell, and
  Maleki}{Warnock-Smith et~al\mbox{.}}{2017}]%
        {warnock2017analysis}
\bibfield{author}{\bibinfo{person}{David Warnock-Smith},
  \bibinfo{person}{John~F O'Connell}, {and} \bibinfo{person}{Mahnaz Maleki}.}
  \bibinfo{year}{2017}\natexlab{}.
\newblock \showarticletitle{An analysis of ongoing trends in airline ancillary
  revenues}.
\newblock \bibinfo{journal}{\emph{Journal of Air Transport Management}}
  \bibinfo{volume}{64} (\bibinfo{year}{2017}), \bibinfo{pages}{42--54}.
\newblock


\bibitem[\protect\citeauthoryear{Woo, Ji, Kono, Tamura, Kuroda, Sugano,
  Yamamoto, Yamashita, and Asama}{Woo et~al\mbox{.}}{2017}]%
        {woo2017lane}
\bibfield{author}{\bibinfo{person}{Hanwool Woo}, \bibinfo{person}{Yonghoon Ji},
  \bibinfo{person}{Hitoshi Kono}, \bibinfo{person}{Yusuke Tamura},
  \bibinfo{person}{Yasuhide Kuroda}, \bibinfo{person}{Takashi Sugano},
  \bibinfo{person}{Yasunori Yamamoto}, \bibinfo{person}{Atsushi Yamashita},
  {and} \bibinfo{person}{Hajime Asama}.} \bibinfo{year}{2017}\natexlab{}.
\newblock \showarticletitle{Lane-change detection based on vehicle-trajectory
  prediction}.
\newblock \bibinfo{journal}{\emph{IEEE Robotics and Automation Letters}}
  \bibinfo{volume}{2}, \bibinfo{number}{2} (\bibinfo{year}{2017}),
  \bibinfo{pages}{1109--1116}.
\newblock


\bibitem[\protect\citeauthoryear{Zhang, Jiang, Liu, and Song}{Zhang
  et~al\mbox{.}}{2020}]%
        {zhang2020spatio}
\bibfield{author}{\bibinfo{person}{Kai Zhang}, \bibinfo{person}{Yushan Jiang},
  \bibinfo{person}{Dahai Liu}, {and} \bibinfo{person}{Houbing Song}.}
  \bibinfo{year}{2020}\natexlab{}.
\newblock \showarticletitle{Spatio-Temporal Data Mining for Aviation Delay
  Prediction}. In \bibinfo{booktitle}{\emph{2020 IEEE 39th International
  Performance Computing and Communications Conference (IPCCC)}}. IEEE,
  \bibinfo{pages}{1--7}.
\newblock


\end{thebibliography}

\end{document}